\documentclass{article}

    \PassOptionsToPackage{numbers, compress}{natbib}
\usepackage[preprint]{neurips_2025}

\usepackage[utf8]{inputenc} % allow utf-8 input
\usepackage[T1]{fontenc}    % use 8-bit T1 fonts
\usepackage{hyperref}       % hyperlinks
\usepackage{url}            % simple URL typesetting
\usepackage{booktabs}       % professional-quality tables
\usepackage{amsfonts}       % blackboard math symbols
\usepackage{nicefrac}       % compact symbols for 1/2, etc.
\usepackage{microtype}      % microtypography
\usepackage{graphicx}
\usepackage{multirow}
\usepackage{amsmath}
\usepackage{array}
\usepackage{multirow}
\usepackage{longtable}
\usepackage{booktabs}
\usepackage{hhline}
\usepackage[table]{xcolor}
\usepackage[normalem]{ulem}
\useunder{\uline}{\ul}{}
\title{\raisebox{-0.5ex}{\includegraphics[height=1.1em]{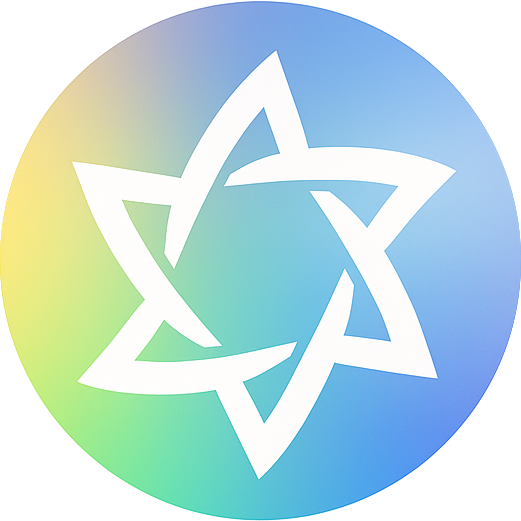}}
OmniGenBench: A Benchmark for Omnipotent Multimodal Generation across 50+ Tasks}

\author{%
  Jiayu Wang\textsuperscript{1,2}\thanks{Equal contribution.},
  Yang Jiao\textsuperscript{1,2}\footnotemark[1],
  Yue Yu\textsuperscript{1,2},
  Tianwen Qian\textsuperscript{3},\\
  \textbf{Shaoxiang Chen\textsuperscript{4},
  Jingjing Chen\textsuperscript{1,2}\thanks{Corresponding author.},
  Yu-Gang Jiang\textsuperscript{1,2}} \\
  \textsuperscript{1}Shanghai Key Lab of Intell. Info. Processing, Fudan University \\
  \textsuperscript{2}Shanghai Collaborative Innovation Center of Intelligent Visual Computing \\
  \textsuperscript{3}School of Computer Science and Technology, East China Normal University \\
  \textsuperscript{4}MiniMax
}

\begin{document}

\maketitle

\begin{abstract}
  Recent breakthroughs in large multimodal models (LMMs), such as the impressive GPT-4o-Native, have demonstrated remarkable proficiency in following general-purpose instructions for image generation. However, current benchmarks often lack the necessary breadth and depth to fully evaluate the diverse capabilities of these models. To overcome this limitation, we introduce \textbf{OmniGenBench}, a novel and comprehensive benchmark meticulously designed to assess the instruction-following abilities of state-of-the-art LMMs across both perception-centric and cognition-centric dimensions. Our OmniGenBench includes 57 diverse sub-tasks grounded in real-world scenarios, systematically categorized according to the specific model capabilities they demand. For rigorous evaluation, we further employ a dual-mode protocol. This protocol utilizes off-the-shelf visual parsing tools for perception-centric tasks and a powerful LLM-based judger for cognition-centric tasks to assess the alignment between generated images and user instructions. Using OmniGenBench, we evaluate mainstream generative models, including prevalent models like GPT-4o, Gemini-2.0-Flash, and Seedream, and provide in-depth comparisons and analyses of their performance. Code and data are available at 
 \href{https://github.com/emilia113/OmniGenBench}{\textit{\textcolor{cyan!100!gray!50!}{OmniGenBench}}}.
  
\end{abstract}

\section{Introduction}
\begin{figure*}
    \centering
    \includegraphics[width=0.9\textwidth]{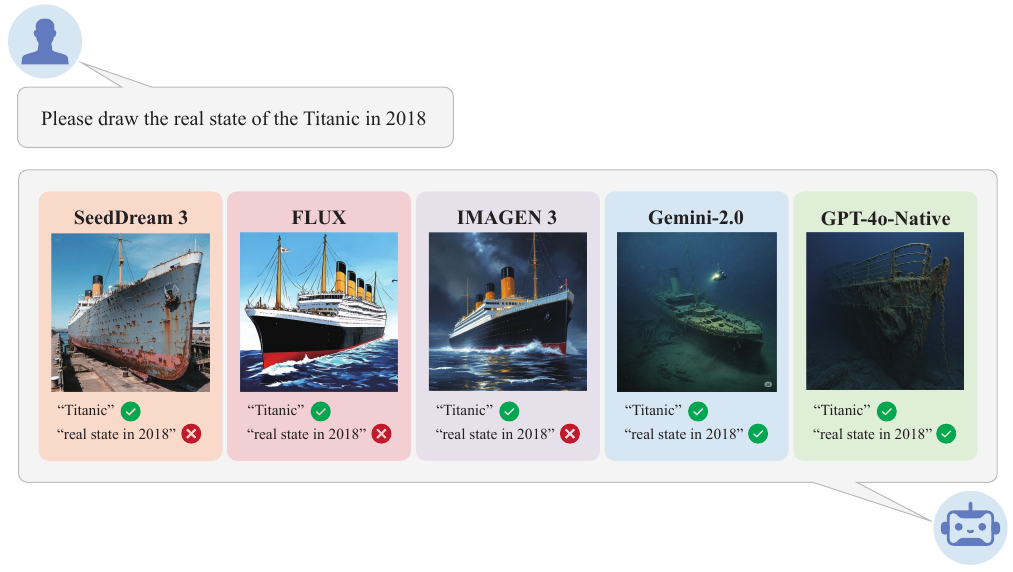}
    \caption{Evaluation of leading generative models in handling the general-purpose instructional task.}
    \label{Comparisonfig}
\end{figure*}

Recently, the emergence of GPT-4o-Native~\cite{hurst2024gpt} has sparked a surge of enthusiasm~\cite{jiao2025unitoken,liu2025step1x,gao2025seedream} for exploring the potential of omnipotent generative models. Reviewing the evolution of the text-to-image generation field, the focus has gradually shifted from the sole pursuit of visual realism to ensuring the consistency between generated visual content and user instructions as conditions. Pioneering works like Stable Diffusion~\cite{rombach2022high} and GLIDE~\cite{nichol2021glide} shed light on high-quality text-to-image generation with the latent diffusion framework, while semantic concepts are typically conveyed through a series of discrete, often handcrafted phrases, known as ``magic prompts", which may limit the expressiveness and controllability of the generation. Driven by the remarkable comprehension capability of MLLMs~\cite{chen2025janus,chen2025sharegpt4v,bai2023qwenvl,jiao2024lumen}, follow-up efforts like DALL-E3~\cite{betker2023improving} and SD3~\cite{esser2024scaling} allow for image generation from coherent natural language prompts, eliminating the reliance on handcrafted prompt engineering. However, as shown in Fig.~\ref{Comparisonfig}, these approaches exhibit persistent difficulties in following general instructions, which hampers their effectiveness in addressing urgent real-world tasks—such as Photoshop (PS) and poster generation. In contrast, GPT-4o-Native and Gemini 2.0-Flash~\cite{team2023gemini} demonstrate remarkable generalization in general-purpose instruction following across diverse domains and modalities, significantly pushing the boundaries of image generation.

% intricate image composition, and reasoning-driven content generation.

% In recent years, text-to-image generation models have undergone a gradual evolution: from early models relying on keyword-based “magic prompts”, to models capable of understanding natural language instructions, and to general-purpose models that can follow complex generation commands and perform a wide range of generation tasks. A leading example in the latest wave of instruction-based generation,  GPT-4o, demonstrates universal generative capabilities. It not only produces rich and high-quality visual output, but also fully understands the underlying goals of instructions, enabling it to carry out open-ended and semantically complex generation tasks exhibiting signs of genuine generative intelligence.To accurately identify and compare the emerging generative capabilities of such models, it is essential to establish an effective evaluation framework.

In line with the evolving generation models, image generation benchmarks~\cite{ghosh2023geneval,huang2025t2i} have also undergone continuous refinement and iteration, with increasing emphasis on evaluating the consistency between generated visual content and user instructions. Early benchmarks, such as COCO~\cite{lin2014microsoft}, primarily focus on evaluating the authenticity of generated image distributions as well as their overall alignment with the corresponding textual inputs. Recently, a stream of benchmarks~\cite{ghosh2023geneval,huang2023t2i,huang2024smartedit,zhao2025envisioning} has emerged to evaluate the instruction-following capabilities of text-to-image generation models from distinctive perspectives. GenEval~\cite{ghosh2023geneval} and T2I-CompBench~\cite{huang2023t2i} underscore assessing the accuracy of basic attributes of generated objects, such as number and spatial position. In contrast, ReasonEdit~\cite{huang2024smartedit} and RISE~\cite{zhao2025envisioning} focus on evaluating the models' ability to respond to reasoning-intensive textual instructions.

While insightful, these benchmarks exhibit limitations in the complexity of their text queries and the breadth of domain coverage. Toward this end, in this paper, we propose \textbf{OmniGenBench}, a novel benchmark encompassing over 50 tasks to comprehensively evaluate the versatility and general capabilities of state-of-the-art large multimodal models. To ensure task diversity and difficulty in our OmniGenBench, we build upon MegaBench~\cite{chen2024mega}, a widely acknowledged all-around benchmark in the multimodal understanding domain, as the basis for principled categorization, and construct text queries via a reverse-engineering process. Specifically, we first manually classify each task into one of two categories: reversible or irreversible\footnote{For specific features of both reversible and irreversible tasks, please refer to Fig.~\ref{fig:data_curation} and appendix for details.}. Then, for each reversible task, we design a tailored system prompt to enable GPT-4o to automatically generate text queries for image generation purposes. Finally, these auto-generated queries are reviewed and filtered by human annotators to ensure correctness and, more importantly, to retain a high level of challenge. Through the above delicate curation process, our OmniGenBench ultimately provides a comprehensive evaluation of both perception-centric and cognition-centric capabilities. Under these two primary categories, the benchmark comprises 57 sub-tasks, designed to thoroughly assess the generation performance of prevalent models across a wide range of domains. For systematic evaluation, we further cluster these sub-tasks into 6 representative categories, enabling a structured assessment of different dimensions of model capabilities.
% 这段还要写的东西：MegaBench来背书query的复杂性 + 逆向理解query时候的筛选以及改写的准则/精神 --> 经过上面这一通操作，最终我们的benchmark是如何呈现的。（一级分类 + 二级分类，把分类的精神/特点说的详细一些）

To enable precise performance assessment, we design separate evaluation protocols tailored to the characteristics of the perception-centric and cognition-centric task categories. For perception-centric tasks, given that current multimodal large models~\cite{chen2025janus,chen2025sharegpt4v,bai2023qwenvl,jiao2024lumen} still show limited proficiency in fundamental visual understanding, such as dense detection and depth estimation, we adopt an automated evaluation approach inspired by GenEval~\cite{ghosh2023geneval}. Specifically, we leverage off-the-shelf scene parsers~\cite{ranftl2021vision,cheng2022masked} to extract object information and assess the correctness of the generated images. For cognition-centric tasks, we employ the LLM-as-a-Judge~\cite{gu2024survey} paradigm, as used in prior studies~\cite{niu2025wise,zhao2025envisioning}, to assess the quality and instruction alignment of the generated images. It is worth mentioning that each sub-task may prioritize distinct key reference points for evaluation. As shown in Fig. 1, whether the Titanic had already sunk is a key criterion for evaluating the accuracy of images showing its actual state in 2018. Therefore, we construct a tailored evaluation prompt for every sub-task, specifying the key reference points that should guide the assessment process. Empirical validation shows that our evaluation protocols align closely with human assessments, providing a robust foundation for automated and reproducible evaluation in future text-to-image generation research. Based on OmniGenBench, we conducted a systematic evaluation of current mainstream generative models. Our experimental results indicate that GPT-4o-Native substantially surpasses other contemporary models in handling tasks that demand adherence to complex reasoning and the incorporation of extensive world knowledge. We hope these results can suggest plausible avenues for further optimization efforts in the research community.
% 下面这段要写的东西：对于perception和cognition具体的测评流程，并且这个测评流程和human evaluation展示出了高度的一致性。而且我们对多少个sota模型展开了系统的测评，测评发现xxx，这说明了xxx，为后续的研究指明了方向

In summary, our contributions are three-fold: 
(1) We propose OmniGenBench, a novel and comprehensive benchmark comprising 57 sub-tasks across perception-centric and cognition-centric dimensions. It is designed to systematically assess the instruction-following capabilities of state-of-the-art generative models in a wide range of real-world domains and task types.
(2) We introduce a task-specific query generation and evaluation framework, which leverages system-level prompting and human filtering for constructing high-quality prompts, and employs a combination of automated metrics and LLM-as-a-judge protocols to ensure reliable, reproducible performance assessment aligned with human judgment.
(3) Extensive experiments show that GPT-4o significantly outperforms other mainstream multimodal models, particularly on complex reasoning tasks and those requiring broad world knowledge, highlighting its superior generalization and instruction-following abilities.

\section{Related Work}

\subsection{Text-to-Image Generation Models}
Text-to-image (T2I) generation aims to synthesize high-quality, diverse images from textual inputs, and can be broadly categorized into two paradigms: dedicated T2I models and unified multimodal models.
Dedicated T2I models have mainly evolved along two architectural lines: Autoregressive~\cite{fan2024fluid,sun2024autoregressive,tian2024visual,chen2020generative,han2024infinity} and Diffusion-based~\cite{ho2020denoising,song2020denoising,ramesh2022hierarchical,saharia2022photorealistic,nichol2021glide}. While autoregressive models treat image generation as a sequential prediction task, they often suffer from high computational costs and limited visual fidelity. In contrast, diffusion models, notably GLIDE~\cite{nichol2021glide} and Latent Diffusion Models~\cite{rombach2022high}—have become the dominant approach due to their superior image quality and efficiency. In parallel, the goal of unified multimodal models is to create general-purpose systems that process text and images and support cross-modal understanding and generation. Recent models like Transfusion~\cite{zhou2024transfusion}, Show-o~\cite{xie2024show}, and D-DiT~\cite{li2024dual} unify the denoising mechanism and autoregressive modeling within a single architecture.

\subsection{Image Generation Benchmarks}
In the context of text-to-image generation, numerous benchmarks~\cite{huang2023t2i,huang2025t2i,zhao2025envisioning,wu2024conceptmix,peng2024dreambench++} have been developed to reflect typical generation scenarios. Notable examples include DPG-Bench~\cite{hu2024ella}, which evaluates dense prompt following, T2I-CompBench~\cite{huang2023t2i}, which targets compositional generation with multiple attributes, and GenEval~\cite{ghosh2023geneval}, which assesses object-level factors such as co-occurrence, spatial layout, and color. However, these benchmarks remain limited in scope and are insufficient for evaluating the general-purpose capabilities of the latest generation of models. Therefore, in this paper, we advance the systematic evaluation of multimodal instruction-following capabilities through OmniGenBench, a comprehensive benchmark encompassing 50+ tasks across diverse real-world scenarios.

\section{OmniGenBench}

We present OmniGenBench, a comprehensive benchmark for assessing the omnipotent instruction-following capabilities of mainstream open-source and commercial models across diverse image generation tasks. The following sections elaborate on the task coverage, data construction pipeline, and evaluation methodology.
% It covers a wide spectrum of tasks ranging from visually grounded content generation to cognitively demanding scenarios involving reasoning, abstraction, and world knowledge. Tasks are categorized into perception-centric and cognition-centric types, enabling a comprehensive assessment of models’ ability to handle both explicit descriptions and implicit visual reasoning.

\subsection{Task Coverage}

\begin{figure*}
    \centering
    \includegraphics[width=1.0\textwidth]{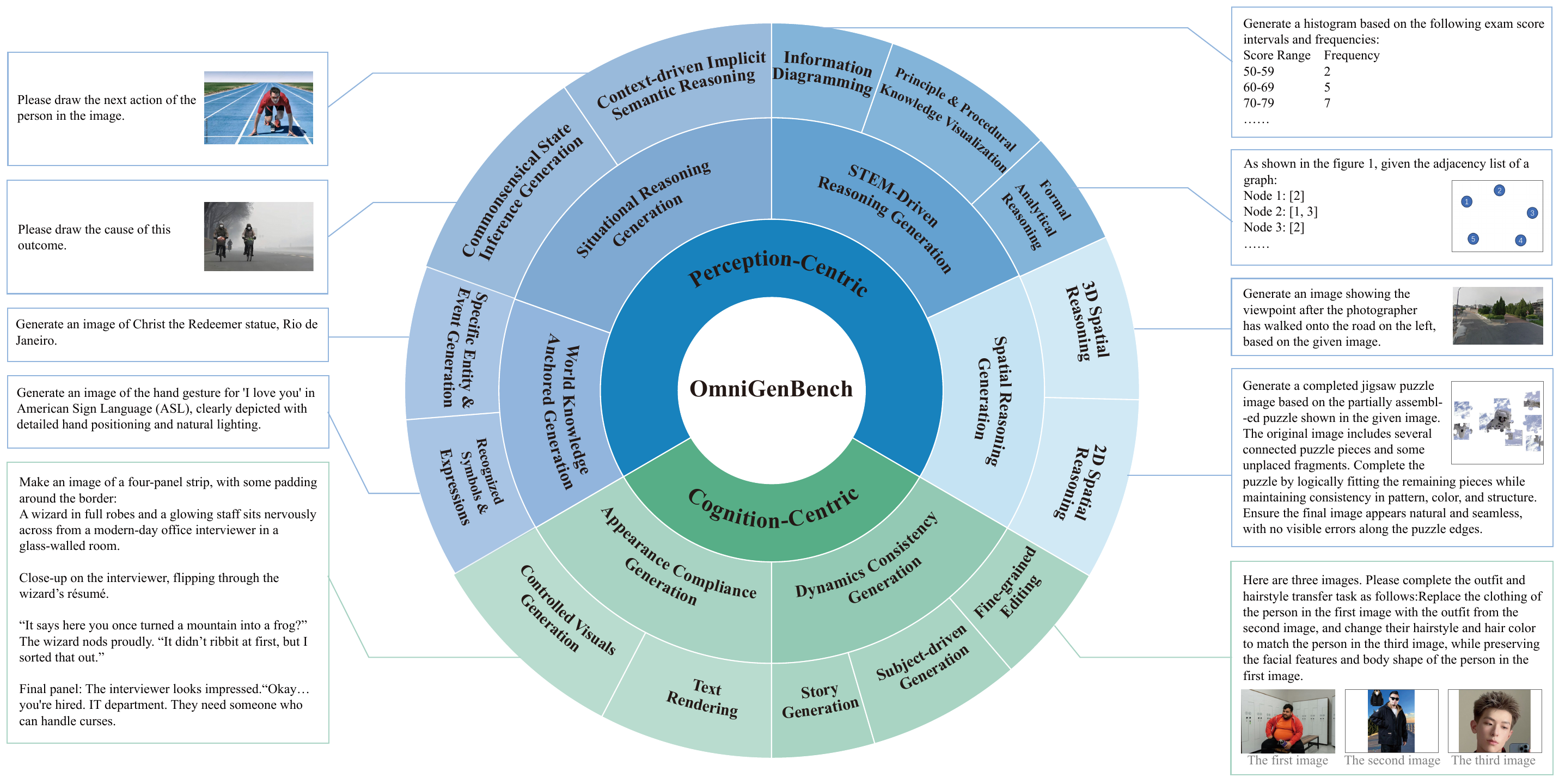}
    \caption{The proposed taxonomy of OmniGenBench, accompanied by representative tasks sampled from across different categories.}
    \label{fig:main}
\end{figure*}

Our benchmark adopts a comprehensive and structured approach to evaluating multimodal generation models. Starting from 57 carefully designed sub-tasks that reflect the diverse demands of real-world generation scenarios, we further cluster them into six major capability dimensions. These dimensions span a broad spectrum from perception-oriented tasks that assess low-level visual fidelity (e.g., appearance consistency) to cognition-oriented challenges that probe high-level reasoning skills (e.g., STEM-driven reasoning). Together, they form an integrated evaluation framework that captures both the surface-level and conceptual strengths of modern generative systems (see Fig.~\ref{fig:main}).

\subsubsection{Perception-Centric Generation}

Perception-Centric Generation encompasses tasks where visual content is directly derived from instructions, without requiring additional reasoning. 
% These tasks evaluate a model’s ability to accurately map basic entity concepts to visual representations.
Within this category, we further divide all tasks into \textit{Appearance Compliance Generation} and \textit{Dynamics Consistency Generation}, based on whether constraints on visual details are considered during evaluation.

\textbf{Appearance Compliance Generation} focuses on generating images based on textual descriptions that specify fundamental object-level attributes—such as quantity, spatial relationships, and attribute bindings. Unlike prior benchmarks like GenEval~\cite{ghosh2023geneval} and T2I-Compbench~\cite{huang2023t2i}, which typically involve only a small number of objects, our benchmark significantly increases the number of objects and their relationships described in each prompt. This design introduces greater compositional complexity and places a much higher demand on the model’s ability to accurately follow instructions and render detailed visual elements. As a result, this task reveals clear limitations in current models: with the exception of GPT-4o, most models struggle notably in maintaining appearance compliance under such high object density and attribute complexity.

\textbf{Dynamics Consistency Generation} assesses a model’s ability to maintain visual coherence across dynamically changing contexts. This category includes two key settings: (1) conditional generation based on a given image and an accompanying textual instruction, where the model must generate a modified image while preserving the essential visual characteristics of the original; and (2) multi-panel generation from a long textual description, where the model needs to ensure visual consistency across sequentially generated images. Both settings challenge the model to retain core visual details, such as object appearance, color, or spatial arrangement, either from the input image or from previous outputs.

\begin{figure*}
    \centering
    \includegraphics[width=1.0\textwidth]{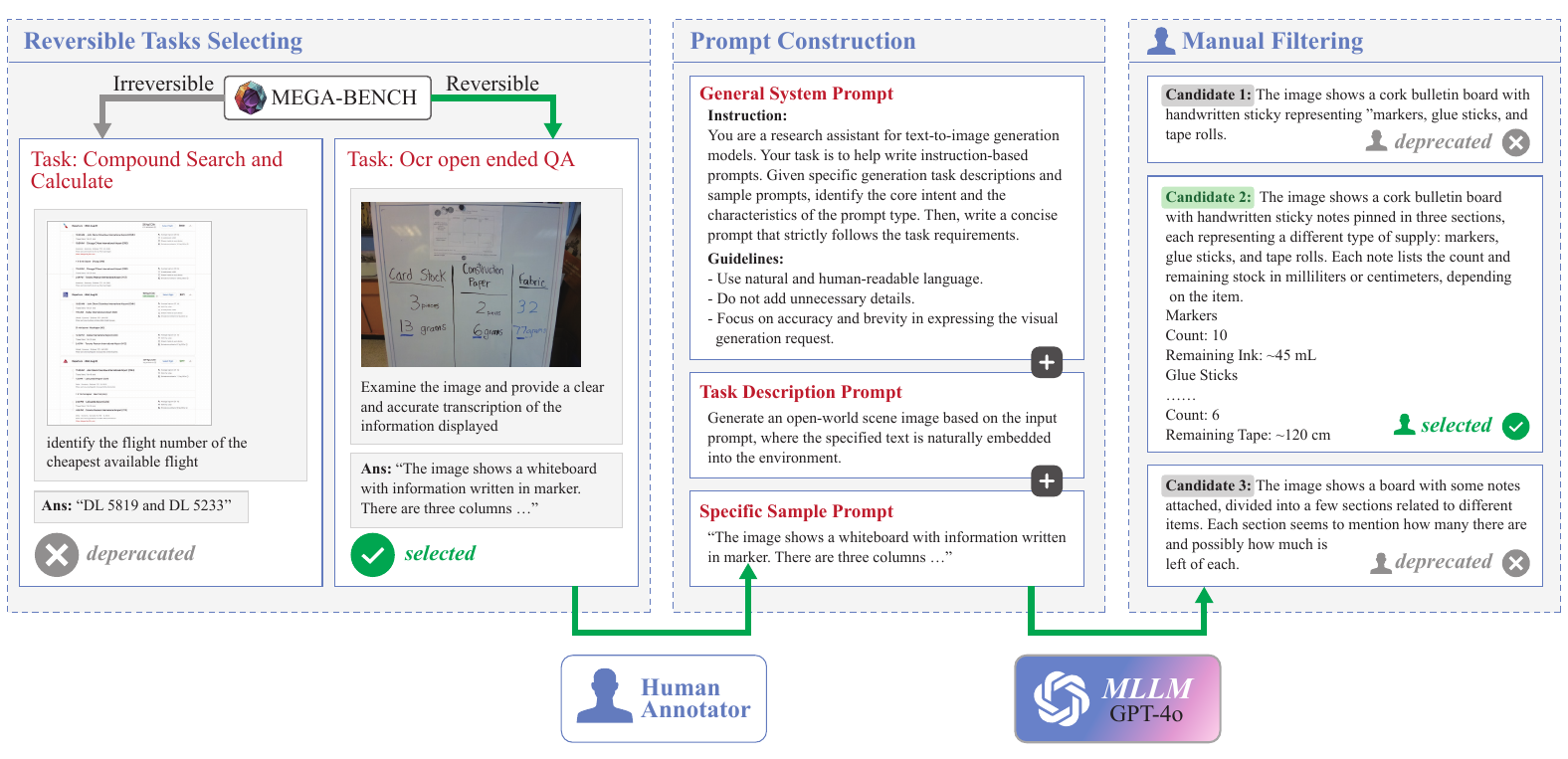}
    \caption{The data curation pipeline of our OmniGenBench.}
    \label{fig:data_curation}
\end{figure*}

\subsubsection{Cognition-Centric Generation}

Cognition-Centric Generation includes tasks that involve inferring visual targets based on prior knowledge and contextual information provided in the instruction, rather than relying solely on literal descriptions. This category comprises four representative types: \textit{World Knowledge Anchored Generation}, \textit{Situational Reasoning}, \textit{Spatial Reasoning}, and \textit{STEM-Driven Reasoning}, based on the distinct reasoning pathways and types of cognitive engagement required in each task.

\textbf{World Knowledge Anchored Generation} evaluates the model’s ability to generate images based on worldwide concepts, rather than common objects with explicit visual features. Unlike perception-centric tasks, this category involves concepts of specialized terms such as \textit{"Statue of Jesus in Brazil"}. To ensure comprehensive coverage, it includes 12 fine-grained sub-domains, ranging from iconic landmarks to logographic script and sign language gestures.

\textbf{Situational Reasoning Generation} requires the model to be placed within an objective scenario and generate images by reasoning about what is likely to happen next, based on its understanding of the surrounding context and real-world principles. Specifically, this category focuses on evaluating the model’s ability to (1) comprehend contextual scenarios and (2) reason based on general principles such as temporality and causality. A typical example within this category is \textit{``Illustrating the effect of a cigarette butt thrown into a forest"}.

% refers to inferring based on the input semantic and visual context to generate images that are consistent with situational semantics or the cognitive principles of common sense.This ability can be divided into the following two categories:

% Context-Driven Implicit Semantic Reasoning Generation: In these tasks, a specific type of generation target is explicitly specified, and the model is required to generate an image that fulfills the task's objective based on the situational information provided in the instructions.For example, the model is required to generate relevant images based on a conversation, which involves inferring the implicit visual elements from the dialogue.These tasks assess the model's ability to understand complex scenarios and infer implicit information.
% Commonsensical State Inference Generation: These tasks leverage common sense to infer the state transitions of objects within the described scenario.The task can be subdivided into two types tasks:

% Time-based Reasoning: This task requires the model to predict object states over time based on objective principles (e.g.,draw the condition of a fresh apple in three months), assessing its ability to infer temporal changes.

% Causal Reasoning: This task requires the model to understand the impact of external forces on objects and predict the outcomes of their influence (e.g., illustrating the effect of a cigarette butt thrown into a forest), assessing the model's understanding of causal relationships.

\textbf{Spatial Reasoning Generation} evaluates a model’s capacity to comprehend and manipulate spatial relationships across 2D and 3D domains—an essential step toward functioning as a world simulator. Toward this end, we specifically devise this category for evaluation. Specifically, we assess the model’s ability to generate 2D and 3D layouts, as well as its capacity for converting between 2D and 3D perspectives.
% tasks assess a model’s ability to understand and reason about spatial relationships within two-dimensional(2D) and three-dimensional(3D) spaces. 

% 3D Spatial Reasoning requires the model to infer the structure and perspective of objects within 3D space, based on its understanding of complex 3D structures and the spatial relationships between objects. The model is expected to generate images that are consistent with 3D spatial information and adhere to the physical laws of the real world.Tasks in this category include:
% 2D-to-3D Conversion: Converting 2D images or representations into 3D space.
% 3D Perspective Transformation: Generating or adjusting views based on different 3D perspectives.
% Generating Complex 3D Scenes: Creating images that depict intricate 3D spatial layouts and physical phenomena.

% Spatial Reasoning involves tasks that evaluate the model’s ability to understand geometric properties and relationships in two-dimensional spaces, such as the relative sizes and positions of objects. These tasks include solving 2D puzzles and generating images based on geometric descriptions, among others.

% \textbf{Abstract Knowledge and Logic Visualization}
\textbf{STEM-Driven Reasoning Generation} evaluates the model's generation capability across science, technology, engineering and mathematics (STEM) fields.
This is inspired by the robust reasoning abilities exhibited by state-of-the-art MLLMs in addressing STEM tasks. More specifically, these tasks involve performing reasoning akin to that of MLLMs, followed by the visualization of both the intermediate steps and the final logical conclusions.

Due to space constraints, Fig.~\ref{fig:main} showcases a subset of representative tasks. A complete list and detailed presentation of all 57 sub-tasks can be found in the appendix.

\subsection{Data Construction}
As shown in Fig.~\ref{fig:data_curation}, we leverage tasks from MegaBench for their versatility and the challenges they present. After manually categorizing all tasks into reversible and irreversible types, we instruct human annotators to construct image generation requests for each reversible task by combining the rephrased VQA question with its corresponding answer. Subsequently, with the resulting image generation request as a few-shot demonstration, we further prompt GPT-4o to automatically generate more diverse requests. Finally, to guarantee the quality and difficulty of our benchmark, we further employ three human annotators to assess the GPT-generated image generation requests, and retain only the examples that receive consistent positive judgments from all three annotators.
% Given the wide variety of instruction types, we designed a controlled prompt generation pipeline to ensure quality. We first crafted task descriptions and corresponding positive and negative examples to guide large language models (e.g., GPT) in generating candidate prompts for each task type. These prompts were then batch-generated, rigorously filtered, and paired with relevant conditioning images selected from public datasets and online sources.

% To ensure prompt quality and validity, we conducted thorough checks on factual soundness and task relevance. For prompts involving factual concepts, we employed GPT-4o to verify that the instructions were executable and aligned with real-world knowledge. In addition, we implemented a dual cross-validation process among annotators to ensure that prompt selection met task requirements and maintained high overall data quality.

\subsection{Evaluation Protocol}

\textbf{Dual-mode Evaluation.}\quad
Considering the general challenge of evaluating image generation quality across diverse scenarios, we adopt a dual-mode evaluation strategy in OmniGenBench. Specifically, for perception-centric tasks, where evaluation emphasizes low-level visual fidelity and the tasks typically correspond to well-established benchmarks (e.g., object attribute generation, image editing), we directly follow the evaluation protocols proposed in previous work~\cite{wang2024evolving,ghosh2023geneval}.

\begin{figure*}
    \centering
    \includegraphics[width=1.0\textwidth]{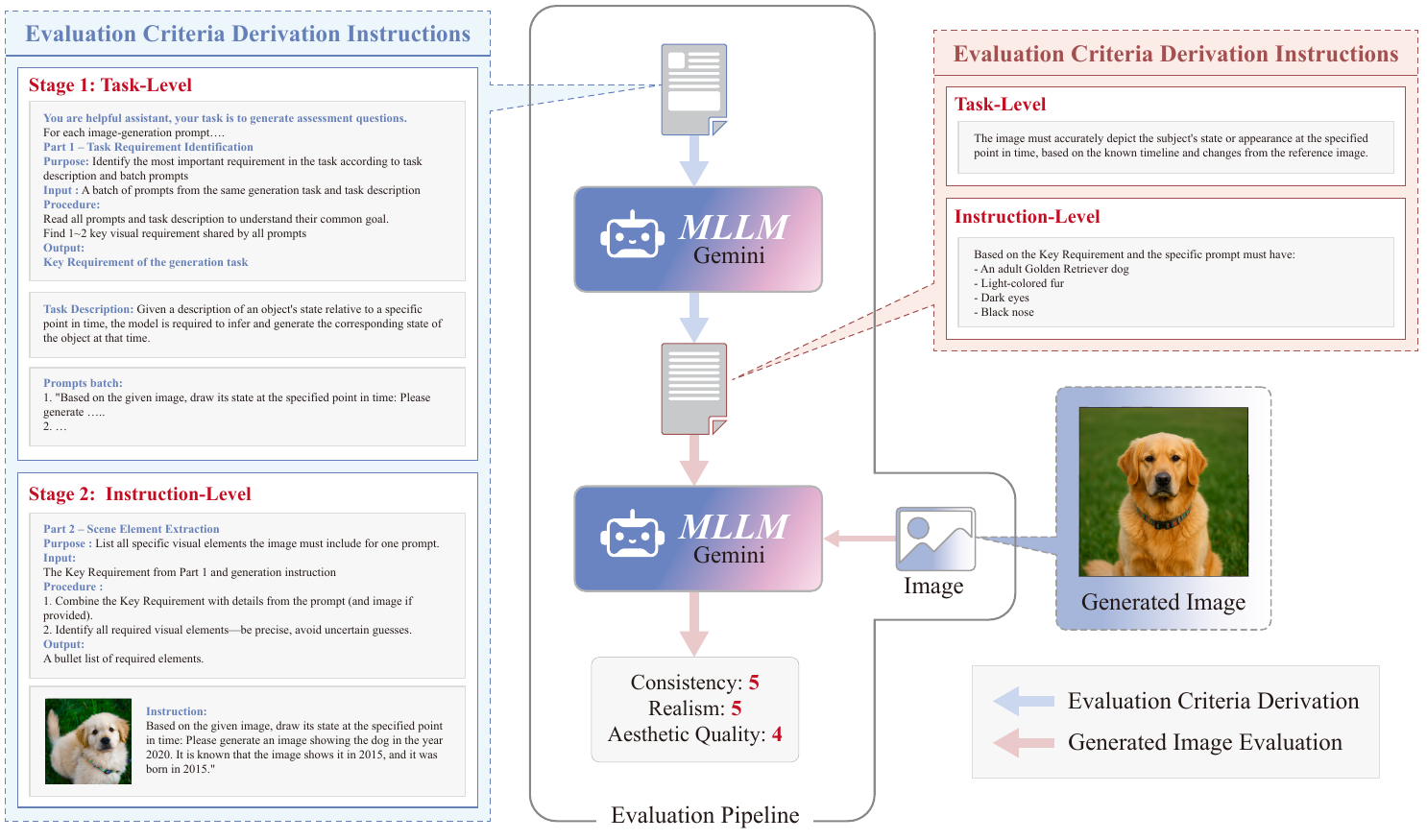}
    \caption{Illustration of the evaluation pipeline of our OmniGenBench.}
    \label{eval_pipeline}
\end{figure*}

In contrast, for cognition-centric tasks, where cognitive engagement varies significantly across different tasks, we follow prior work~\cite{zhao2025envisioning} and adopt the MLLM-as-a-Judge paradigm. To ensure that the MLLM can accurately assess the quality of image generation tailored to each specific task, we design customized evaluation prompts for each task. These prompts include task-specific key reference points that guide the model’s evaluation focus. To efficiently generate these customized evaluation prompts, we develop an automated evaluation pipeline, as illustrated in Fig.~\ref{eval_pipeline}. Concretely, we first embed both the task description and the current generation instruction into the system prompt, and then prompt Gemini-2.5-Pro to generate the task-specific evaluation criteria. The resulting criteria, together with the generated image, are subsequently fed into Gemini-2.5-Pro to obtain the final quality score.

\begin{figure*}
    \centering
    \includegraphics[width=0.75\textwidth]{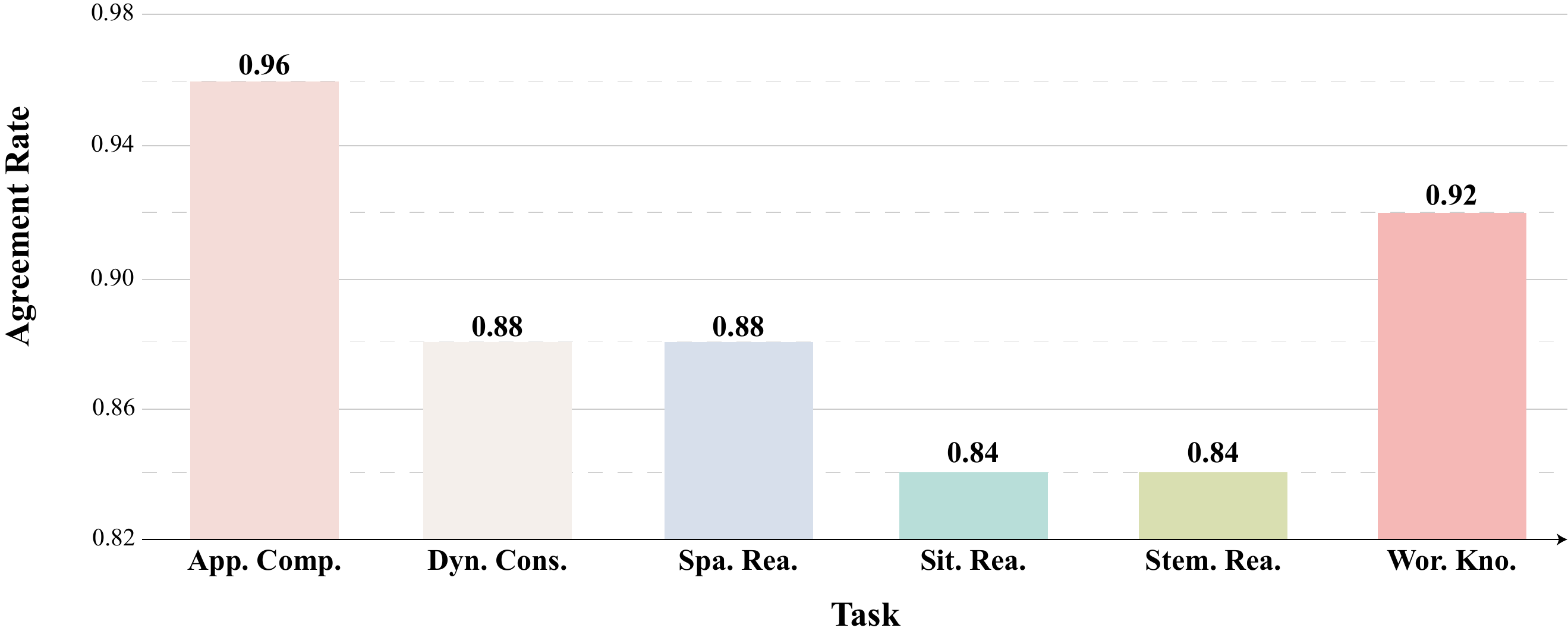}
    \caption{Results of agreement rate between our OmniScore and human evaluation.}
    \label{fig:human_agree}
\end{figure*}
\textbf{Evaluation Metrics.} \quad
We propose a multi-dimensional evaluation framework to rigorously assess the quality of generated images, focusing on four core aspects: Consistency, Realism, Aesthetics Quality, and combining them to obtain the evaluation metric we termed as OmniScore (as illustrated in Figure~\ref{eval_pipeline}). Specifically, consistency evaluates how accurately and completely the generated image reflects the intended meaning of the instruction, based on instruction-specific tailored criteria. Realism measures the visual plausibility or photorealistic quality of the image, assessing whether the image appears credible in the context of real-world scenes. Aesthetics Quality assesses the overall visual appeal of the image, including aspects such as composition, color harmony, and stylistic coherence.

Overall, OmniScore serves as the central metric in our evaluation framework, comprehensively highlighting the model’s general-purpose generation capabilities. It is computed using a weighted formula:
\begin{equation}
    \text{OmniScore} = 0.8 \times \text{Consistency} + 0.1 \times \text{Realism} + 0.1 \times \text{Aesthetics\ Quality}
\end{equation}
This weighting scheme prioritizes adherence to the instruction while incorporating realism and aesthetic quality to ensure holistic image quality. A higher OmniScore indicates stronger generalization capability of the model across diverse generation tasks.

% Please add the following required packages to your document preamble:
% \usepackage{multirow}
% \usepackage[normalem]{ulem}
% \useunder{\uline}{\ul}{}
\begin{table}[t]
\centering
\caption{Evaluation results of mainstream text-to-image generation models on the OmniGenBench. We highlight the best and the second best performances by \textbf{bold} and \uline{underline}, respectively.}
\label{tab:main}
\scalebox{0.9}{
\begin{tabular}{lcccccc}
\toprule
\multicolumn{1}{l|}{\multirow{2}{*}{Model}} & \multicolumn{2}{c|}{Perception-Centric}            & \multicolumn{4}{c}{Cognition-Centric}                         \\ \cmidrule{2-7} 
\multicolumn{1}{l|}{}                       & App. Comp.    & \multicolumn{1}{c|}{Dyn. Cons.}    & Spa. Rea.     & Sit. Rea.     & STEM Rea.     & Wor. Kno.     \\ \midrule
\multicolumn{7}{c}{\textit{Open-source Models}}                                                                                                                  \\ \midrule
\multicolumn{1}{l|}{SD1.5~\cite{rombach2022high}}                  & 35.3          & \multicolumn{1}{c|}{39.2}          & 39.0          & 35.1          & 32.5          & 57.7          \\
\multicolumn{1}{l|}{SDXL~\cite{podell2023sdxl}}                   & 44.5          & \multicolumn{1}{c|}{43.4}          & 57.3          & 50.4          & 43.1          & 70.2          \\ \midrule
\multicolumn{7}{c}{\textit{Close-source Models}}                                                                                                                 \\ \midrule
\multicolumn{1}{l|}{SD3~\cite{esser2024scaling}}                    & 53.8          & \multicolumn{1}{c|}{49.5}          & -          & 29.0          & 46.7          & 75.8          \\
\multicolumn{1}{l|}{FLUX1.1~\cite{flux11pro2024}}               & 58.4          & \multicolumn{1}{c|}{52.7}          & 57.4          & 42.5          & 47.9          & 77.3          \\
\multicolumn{1}{l|}{Seedream2.1~\cite{gong2025seedream}}           & 52.7          & \multicolumn{1}{c|}{50.4}          & 45.6          & 38.9          & 45.0          & 77.0          \\
\multicolumn{1}{l|}{Seedream3~\cite{gao2025seedream}}             & 72.1          & \multicolumn{1}{c|}{62.0}          & -             & 52.0          & 52.5          & 82.7          \\
\multicolumn{1}{l|}{Imagen3~\cite{baldridge2024imagen}}               & 58.8          & \multicolumn{1}{c|}{61.5}          & -             & 42.7          & 52.3          & \textbf{85.1} \\
\multicolumn{1}{l|}{Gemini-2.0~\cite{team2023gemini}}             & {\ul 72.4}    & \multicolumn{1}{c|}{{\ul 66.5}}    & {\ul 81.4}    & {\ul 81.4}    & {\ul 56.7}    & 84.0    \\
\multicolumn{1}{l|}{GPT-4o-Native~\cite{hurst2024gpt}}          & \textbf{82.9} & \multicolumn{1}{c|}{\textbf{92.5}} & \textbf{82.8} & \textbf{82.5} & \textbf{63.7} & {\ul 85.0} \\ \bottomrule
\end{tabular}}
\end{table}
   
\section{Experiments}
This section is organized as follows. We begin with a comprehensive evaluation of mainstream text-to-image generation models on our proposed OmniGenBench. Next, we assess the alignment between our automated metric, OmniScore, and human evaluations. Finally, by qualitatively comparing model outputs on representative hard cases, we identify performance bottlenecks in existing models relative to GPT-4o-Native, thereby uncovering critical areas for future advancement.

\begin{figure*}
    \centering
    \includegraphics[width=1.0\textwidth]{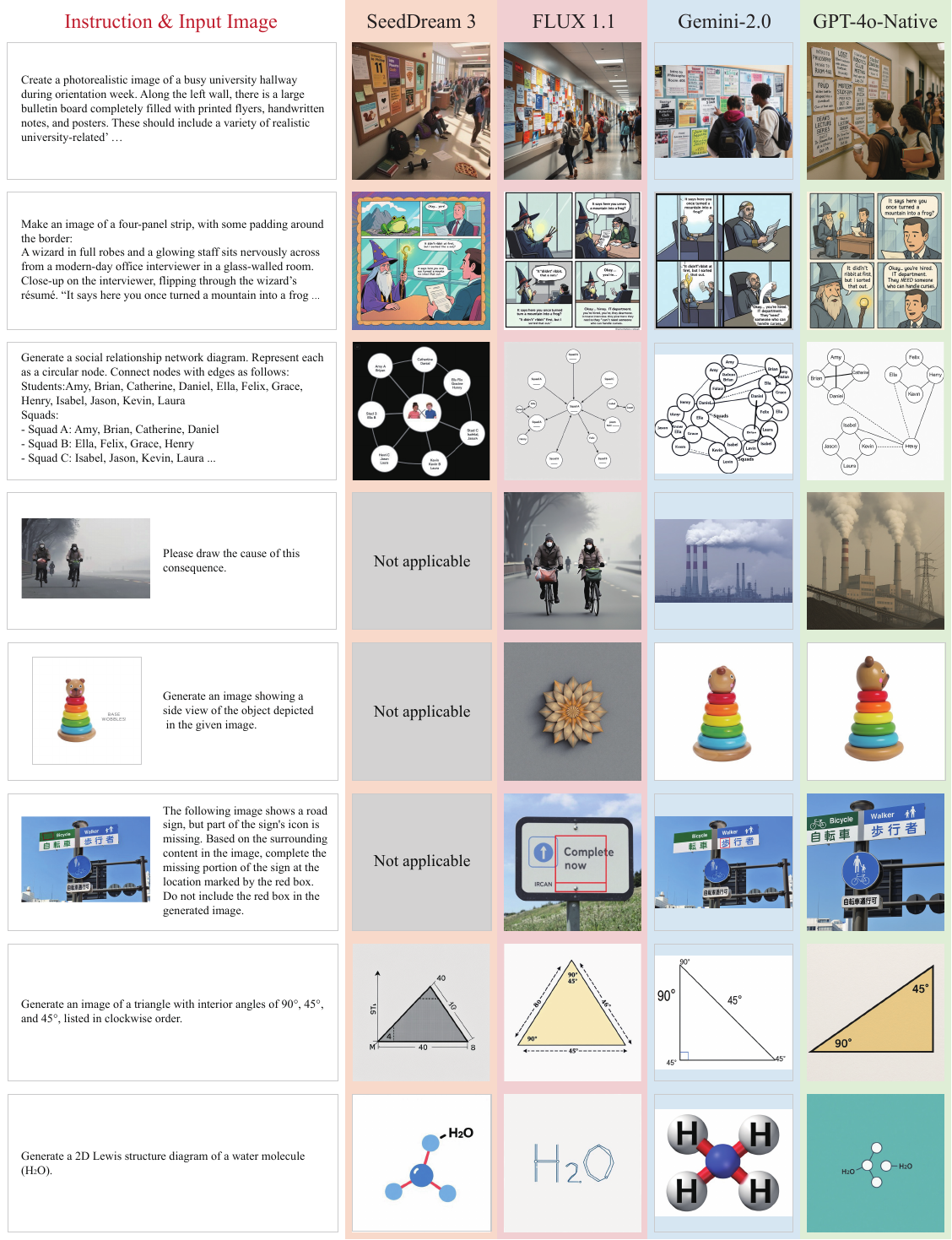}
    \caption{Qualitative comparison of four leading text-to-image generation methods on the OmniGenBench. The gray tiles labeled “Not applicable” indicate unsupported input types for the corresponding model.}
    \label{fig:qual}
\end{figure*}

\subsection{Evaluation of Mainstream Approaches}
We evaluate a range of mainstream text-to-image generation models on the OmniGenBench, by assessing both perception-centric and cognition-centric capabilities across six major categories. The results are summarized in Table~\ref{tab:main}, where we have the following observations.
\textbf{(1)} Overall, GPT-4o-Native demonstrates state-of-the-art performance across all dimensions, establishing itself as the most capable model in both perception and cognition tasks. It achieves the highest scores across almost all categories, indicating a well-rounded and highly advanced general-purpose text-to-image generation ability.
\textbf{(2)} Gemini-2.0 ranks as the second-best model overall. It exhibits strong cognition-oriented generation capabilities, with especially high performance in Spatial Reasoning (81.4) and Situational Reasoning (81.4), and competitive scores in World Knowledge (84.0). However, its performance on perception-oriented tasks lags behind GPT-4o-Native.
\textbf{(3)} Among the other close-sourced models, Seedream3 stands out in terms of perception performance, particularly in Appearance Compliance (72.1), surpassing most other closed models except Gemini-2.0 and GPT-4o-Native. However, its cognitive reasoning capabilities are relatively weaker, especially in STEM Reasoning (52.5) and Situational Reasoning (52.0), which limits its overall versatility.
\textbf{(4)} In contrast, open-source models (e.g., SD1.5 and SDXL) show substantially lower performance across nearly all metrics. For instance, SD1.5 achieves only 35.3 in Appearance Compliance and 32.5 in STEM Reasoning, highlighting a significant gap in both perceptual and cognitive capabilities compared to their closed-source counterparts. SDXL offers moderate improvements but still remains far behind the best-performing closed-source models. This discrepancy may stem from the significantly larger parameter scales and more extensive training datasets typically available to proprietary models, enabling more robust generalization across diverse generation tasks.

\subsection{Agreement with Human Evaluation}
In this section, we assess the consistency between our benchmark evaluation metric, OmniScore, and human evaluation. Specifically, we sample a subset from six categories within our dataset, comparing the evaluations produced by OmniScore with those provided by human annotators. To minimize bias, three annotators were tasked with independently determining whether the generated images aligned with the provided textual instructions. Only the samples for which all three annotators agreed were retained for further analysis, and their results were cross-validated with the corresponding OmniScore assessments.
Finally, the agreement between OmniScore and human evaluation was calculated for each category independently. Results in Fig.~\ref{fig:human_agree} indicate a high level of consistency, with OmniScore demonstrating strong alignment with human judgments across the diverse categories in the dataset. This suggests that OmniScore is a reliable and effective metric for evaluating image generation quality in a manner that closely mirrors human judgement.

\subsection{In-depth Qualitative Comparison}
Through a comprehensive examination of images generated by leading methods across all sub-tasks in OmniGenBench, we select a set of challenging cases to intuitively compare model capabilities across multiple dimensions. From Fig.~\ref{fig:qual}, we observe that GPT-4o-Native excels in the following key aspects: (1) \textbf{Text Generation.} As shown in $\#1$, GPT-4o-Native can generate words more vivid and clearer than other models. (2) \textbf{ID-Preserving.} As demonstrated in $\#2$, GPT-4o-Native maintains strong appearance consistency for the same person across different panels, while other models perform poorly in comparison. (3) \textbf{Structured Knowledge Representation.} As shown in $\#3$, GPT-4o-Native is the only model capable of both summarizing the information correctly and producing an accurate graph structure. (4) \textbf{Commonsense Reasoning.} As shown in $\#4$, GPT-4o-Native can deduce a plausible source of the pollution. Although Gemini-2.0 can make similar assumptions, the background style is inconsistent with the input image. 
(5) \textbf{Spatial Reasoning.} As shown in $\#5$, GPT-4o-Native can comprehend the 3D geometric structure of the object, and thus make accurate inferences about novel viewpoints. (6) \textbf{In-Context Reasoning.} As shown in $\#6$, GPT-4o-Native deduces the presence of a bicycle logo in the marked area via cross-referencing the pattern of surrounding regions.

While GPT-4o-Native excels in many areas, we also document its limitations through specific failure cases. As shown in $\#7,\#8$, all four leading methods fail to generate elements requiring strong professional knowledge.
% \textbf{(1)} As shown in the results of the first row, compared to other models, GPT-4o-Native demonstrates a \textbf{stronger text generation capability, as evidenced by its ability to produce clearly legible text in real-world spatia   l contexts. \textbf{(2)} In terms of ID-preserving, GPT-4o-Native also presen

\section{Conclusion}
In this paper, we presented OmniGenBench, a comprehensive benchmark for evaluating the instruction-following abilities of LMMs across perception and cognition dimensions. Through 57 diverse tasks and a dual-mode evaluation protocol, our results highlight key strengths and limitations of current models compared with GPT-4o-Native. OmniGenBench offers a solid foundation for guiding future improvements in multimodal generation.
\bibliographystyle{plainnat}
\bibliography{neurips_2025}

\clearpage
\appendix

\section{Detailed Task Taxonomy of OmniGenBench}
The table~\ref{tab:detailed_taxonomy} presents the detailed structure of the OmniGenBench task taxonomy. At the first level, tasks are categorized into Cognition-centric and Perception-centric groups based on their intrinsic nature. At the second level, they are further organized into six major capability dimensions, with all specific tasks under each dimension explicitly listed. 

To improve clarity, the table is segmented into distinct regions according to the first- and second-level categories. The level 3 and level 4 task names are displayed within each region, and every level 4 task is assigned a unique identifier. If a level 4 task contains finer-grained subcategories, these are clearly indicated within the corresponding entry.

\section{Task Examples and Qualitative Comparison of Generation Models}
For each specific task in our benchmark, we select a representative example to illustrate the task objective. Alongside each example, we present the corresponding outputs generated by mainstream generative models. This side-by-side comparison facilitates direct observation of each model’s strengths and weaknesses in handling diverse generation instructions.The gray tiles labeled “Not applicable” indicate cases where a model either does not support the given input format.

% Task class table begin
% \clearpage
\renewcommand{\arraystretch}{1.24}  % 默认是 1，1.5 表示行高增加 50%
\begin{longtable}[c]{|>{\hspace{2pt}}p{1cm}<{\hspace{2pt}}|>{\hspace{4pt}}p{4cm}<{\hspace{4pt}}|>{\hspace{4pt}}p{4cm}<{\hspace{4pt}}|}

\caption{Multi-Level Task Taxonomy of OmniGenBench.}\label{tab:detailed_taxonomy} \\
\hline
\multicolumn{1}{|c|}{\rule{0pt}{2.1em}\textbf{Task ID}\rule[-1.2em]{0pt}{0pt}} & \multicolumn{1}{c|}{\rule{0pt}{2.1em}\textbf{Level 3 Task}\rule[-1.2em]{0pt}{0pt}}                                                                                         & \multicolumn{1}{c|}{\rule{0pt}{2.1em}\textbf{Level 4 Task}\rule[-1.2em]{0pt}{0pt}}                                                                                                                                                                                                                  \\
% \hline
\hhline{|===|}
% \endfirsthead
%
\endhead
\rowcolor{cyan!10}
\multicolumn{3}{|c|}{\rule{0pt}{1.3em}\textit{\textbf{Cognition-centric}}\rule[-0.8em]{0pt}{0pt}}                                                                                                                                                                                                                                                                                                                                                                                           \\ 
\hhline{|===|}
\multicolumn{3}{|c|}{\rule{0pt}{1.2em} \textbf{World Knowledge Anchored Generation} \rule[-0.5em]{0pt}{0pt}}                                                                                                                                                                                                                                                                                                                                                                                  \\ \hline

\multicolumn{1}{|c|}{1.1}              & \multicolumn{1}{c|}{\multirow{4}{*}{\begin{tabular}[c]{@{}c@{}}Specified Entity \\and Event\\ Generation\end{tabular}}}                   & \multicolumn{1}{l|}{\begin{tabular}[c]{@{}l@{}}Concrete Entity Generation {[}Location{]} (1.1.1)\\ Concrete Entity Generation {[}Natural Entities{]} (1.1.2)\\ Concrete Entity Generation {[}Artificial Entities{]} (1.1.3)\end{tabular}}                        \\ \cline{1-1} \cline{3-3} 
\multicolumn{1}{|c|}{1.2}              & \multicolumn{1}{c|}{}                                                                                                                   & \multicolumn{1}{l|}{\begin{tabular}[c]{@{}l@{}}Societal Role \& Event Generation {[}Social Activities{]} (1.2.1)\\ Societal Role \& Event Generation {[}Historical Event{]} (1.2.2)\\ Societal Role \& Event Generation {[}Social Roles{]} (1.2.3)\end{tabular}} \\ \hline
\multicolumn{1}{|c|}{1.3}              & \multicolumn{1}{c|}{\multirow{2}{*}{\begin{tabular}[c]{@{}c@{}}Recognized\\ Symbols and\\  Expressions\\  Generation\end{tabular}}}     & \multicolumn{1}{l|}{\begin{tabular}[c]{@{}l@{}}Human Pose \& Gesture Generation {[}Hand Gestures{]} (1.3.1)\\ Human Pose \& Gesture Generation {[}Single Person{]} (1.3.2)\\ Human Pose \& Gesture Generation {[}Multiple People{]} (1.3.3)\end{tabular}}        \\ \cline{1-1} \cline{3-3} 
\multicolumn{1}{|c|}{1.4}              & \multicolumn{1}{c|}{}                                                                                                                   & \multicolumn{1}{l|}{\begin{tabular}[c]{@{}l@{}}Symbol and Sign Generation {[}Special Script{]} (1.4.1)\\ Symbol and Sign Generation {[}Brand Logos{]} (1.4.2)\\ Symbol and Sign Generation {[}Rule Signs{]} (1.4.3)\end{tabular}}                                \\ \hline
\multicolumn{3}{|c|}{\rule{0pt}{1.2em} \textbf{Situational Reasoning Generation} \rule[-0.5em]{0pt}{0pt}}                                                                                                                                                                                                                                                                                                                                                                                     \\ \hline
\multicolumn{1}{|c|}{2.1}              & \multicolumn{1}{c|}{\multirow{6}{*}{\begin{tabular}[c]{@{}c@{}}Context-Driven \\ Implicit Semantic \\ Reasoning\end{tabular}}}          & \multicolumn{1}{l|}{Conversation Image Inference}                                                                                                                                                                                                                \\ \cline{1-1} \cline{3-3} 
\multicolumn{1}{|c|}{2.2}              & \multicolumn{1}{c|}{}                                                                                                                   & \multicolumn{1}{l|}{Cross View Consistency Editing}                                                                                                                                                                                                              \\ \cline{1-1} \cline{3-3} 
\multicolumn{1}{|c|}{2.3}              & \multicolumn{1}{c|}{}                                                                                                                   & \multicolumn{1}{l|}{Image Completion Reasoning}                                                                                                                                                                                                                  \\ \cline{1-1} \cline{3-3} 
\multicolumn{1}{|c|}{2.4}              & \multicolumn{1}{c|}{}                                                                                                                   & \multicolumn{1}{l|}{Human Action Prediction}                                                                                                                                                                                                                     \\ \cline{1-1} \cline{3-3} 
\multicolumn{1}{|c|}{2.5}              & \multicolumn{1}{c|}{}                                                                                                                   & \multicolumn{1}{l|}{Reference-guided Structural Image Generation}                                                                                                                                                                                                \\ \cline{1-1} \cline{3-3} 
\multicolumn{1}{|c|}{2.6}              & \multicolumn{1}{c|}{}                                                                                                                   & \multicolumn{1}{l|}{Instruction Translation Generation}                                                                                                                                                                                                          \\ \hline
\multicolumn{1}{|c|}{2.7}              & \multicolumn{1}{c|}{\multirow{4}{*}{\begin{tabular}[c]{@{}c@{}}Commonsensical \\ State Inference\end{tabular}}}           & \multicolumn{1}{l|}{Relative Time Prediction}                                                                                                                                                                                                                    \\ \cline{1-1} \cline{3-3} 
\multicolumn{1}{|c|}{2.8}              & \multicolumn{1}{c|}{}                                                                                                                   & \multicolumn{1}{l|}{Specific Timepoint Prediction}                                                                                                                                                                                                               \\ \cline{1-1} \cline{3-3} 
\multicolumn{1}{|c|}{2.9}              & \multicolumn{1}{c|}{}                                                                                                                   & \multicolumn{1}{l|}{Infer Reason from Outcome}                                                                                                                                                                                                                   \\ \cline{1-1} \cline{3-3} 
\multicolumn{1}{|c|}{2.10}             & \multicolumn{1}{c|}{}                                                                                                                   & \multicolumn{1}{l|}{Infer Outcome according to Visual Textual Reason}                                                                                                                                                                                            \\ \cline{1-1} \cline{3-3} \hline
\multicolumn{1}{|c|}{2.11}             & \multicolumn{1}{c|}{\multirow{1}{*}{\begin{tabular}[c]{@{}c@{}} \end{tabular}}}                                                                                                                   & \multicolumn{1}{l|}{Infer Outcome according to Textual Reason}                                                                                                                                                                                                   \\ \hline
\multicolumn{3}{|c|}{\rule{0pt}{1.2em} \textbf{Spatial Reasoning Generation} \rule[-0.5em]{0pt}{0pt}}                                                                                                                                                                                                                                                                                                                                                                                         \\ \hline
\multicolumn{1}{|c|}{3.1}              & \multicolumn{1}{c|}{\multirow{6}{*}{\begin{tabular}[c]{@{}c@{}}3D Spatial \\ Reasoning\end{tabular}}}                                   & \multicolumn{1}{l|}{2D-to-3D Spatial Scene Generation}                                                                                                                                                                                                           \\ \cline{1-1} \cline{3-3} 
\multicolumn{1}{|c|}{3.2}              & \multicolumn{1}{c|}{}                                                                                                                   & \multicolumn{1}{l|}{Multi-view Composition}                                                                                                                                                                                                                      \\ \cline{1-1} \cline{3-3} 
\multicolumn{1}{|c|}{3.3}              & \multicolumn{1}{c|}{}                                                                                                                   & \multicolumn{1}{l|}{Scene Viewpoint Transformation}                                                                                                                                                                                                              \\ \cline{1-1} \cline{3-3} 
\multicolumn{1}{|c|}{3.4}              & \multicolumn{1}{c|}{}                                                                                                                   & \multicolumn{1}{l|}{Object Viewpoint Transformation}                                                                                                                                                                                                             \\ \cline{1-1} \cline{3-3} 
\multicolumn{1}{|c|}{3.5}              & \multicolumn{1}{c|}{}                                                                                                                   & \multicolumn{1}{l|}{Physically Plausible Scene Generation}                                                                                                                                                                                                       \\ \cline{1-1} \cline{3-3} 
\multicolumn{1}{|c|}{3.6}              & \multicolumn{1}{c|}{}                                                                                                                   & \multicolumn{1}{l|}{2D-3D Hybrid Subject Driven Generation}                                                                                                                                                                                                      \\ \hline
\multicolumn{1}{|c|}{3.7}              & \multicolumn{1}{c|}{\multirow{4}{*}{\begin{tabular}[c]{@{}c@{}}2D Spatial \\ Reasoning\end{tabular}}}                                   & \multicolumn{1}{l|}{2D Puzzle Solving}                                                                                                                                                                                                                           \\ \cline{1-1} \cline{3-3} 
\multicolumn{1}{|c|}{3.8}              & \multicolumn{1}{c|}{}                                                                                                                   & \multicolumn{1}{l|}{Angle to Polygon Construction}                                                                                                                                                                                                               \\ \cline{1-1} \cline{3-3} 
\multicolumn{1}{|c|}{3.9}              & \multicolumn{1}{c|}{}                                                                                                                   & \multicolumn{1}{l|}{Side Length to Polygon Construction}                                                                                                                                                                                                         \\ \cline{1-1} \cline{3-3} 
\multicolumn{1}{|c|}{3.10}            & \multicolumn{1}{c|}{}                                                                                                                   & \multicolumn{1}{l|}{Geometric Structure Drawing}                                                                                                                                                                                                                 \\ \hline
\multicolumn{3}{|c|}{\rule{0pt}{1.2em} \textbf{STEM-Driven Reasoning Generation} \rule[-0.5em]{0pt}{0pt}}                                                                                                                                                                                                                                                                                                                                                                                     \\ \hline
\multicolumn{1}{|c|}{4.1}              & \multicolumn{1}{c|}{\multirow{4}{*}{\begin{tabular}[c]{@{}c@{}}\rule{0pt}{1.0em}Principle and\\ Procedural\\ Knowledge \\ Visualization\rule[-0.3em]{0pt}{0pt}\end{tabular}}} & \multicolumn{1}{l|}{Natural \& Life Sciences Concept and Law Visualization}                                                                                                                                                                                             \\ \cline{1-1} \cline{3-3} 
\multicolumn{1}{|c|}{4.2}              & \multicolumn{1}{c|}{}                                                                                                                   & \multicolumn{1}{l|}{Natural \& Life Sciences Reaction and Law Visualization}                                                                                                                                                                                     \\ \cline{1-1} \cline{3-3} 
\multicolumn{1}{|c|}{4.3}              & \multicolumn{1}{c|}{}                                                                                                                   & \multicolumn{1}{l|}{Mathematical Function Graph Generation}                                                                                                                                                                                                      \\ \cline{1-1} \cline{3-3} 
\multicolumn{1}{|c|}{4.4}              & \multicolumn{1}{c|}{}                                                                                                                   & \multicolumn{1}{l|}{Rules and Procedures Visualization}                                                                                                                                                                                                          \\ \hline
\multicolumn{1}{|c|}{4.5}              & \multicolumn{1}{c|}{\multirow{7}{*}{\begin{tabular}[c]{@{}c@{}}Information \\ Diagramming\end{tabular}}}                                & \multicolumn{1}{l|}{Statistical Chart Generation}                                                                                                                                                                                                                \\ \cline{1-1} \cline{3-3} 
\multicolumn{1}{|c|}{4.6}              & \multicolumn{1}{c|}{}                                                                                                                   & \multicolumn{1}{l|}{Causal Chain Diagramming}                                                                                                                                                                                                                    \\ \cline{1-1} \cline{3-3} 
\multicolumn{1}{|c|}{4.7}              & \multicolumn{1}{c|}{}                                                                                                                   & \multicolumn{1}{l|}{Hierarchical Structure Diagramming}                                                                                                                                                                                                          \\ \cline{1-1} \cline{3-3} 
\multicolumn{1}{|c|}{4.8}              & \multicolumn{1}{c|}{}                                                                                                                   & \multicolumn{1}{l|}{Social Relationship Network Diagramming}                                                                                                                                                                                                     \\ \cline{1-1} \cline{3-3} 
\multicolumn{1}{|c|}{4.9}              & \multicolumn{1}{c|}{}                                                                                                                   & \multicolumn{1}{l|}{Reference-guided Structural Image Generation}                                                                                                                                                                                                \\ \cline{1-1} \cline{3-3} 
\multicolumn{1}{|c|}{4.10}    & \multicolumn{1}{c|}{}                                                                                                                   & \multicolumn{1}{l|}{Timeline Diagramming}                                                                                                                                                                                                                        \\ \cline{1-1} \cline{3-3} 
\multicolumn{1}{|c|}{4.11}             & \multicolumn{1}{c|}{}                                                                                                                   & \multicolumn{1}{l|}{Table Generation}                                                                                                                                                                                                                            \\ \hline
\multicolumn{1}{|c|}{4.12}             & \multicolumn{1}{c|}{\multirow{3}{*}{\begin{tabular}[c]{@{}c@{}}Formal Analytical \\ Reasoning\end{tabular}}}                            & \multicolumn{1}{l|}{Graph Problem Solving}                                                                                                                                                                                                                       \\ \cline{1-1} \cline{3-3} 
\multicolumn{1}{|c|}{4.13}             & \multicolumn{1}{c|}{}                                                                                                                   & \multicolumn{1}{l|}{Mathematical Reasoning and Visualization}                                                                                                                                                                                                    \\ \cline{1-1} \cline{3-3} 
\multicolumn{1}{|c|}{4.14}             & \multicolumn{1}{c|}{}                                                                                                                   & \multicolumn{1}{l|}{Step-by-Step STEM Problem Solving}                                                                                                                                                                                                           \\ 
% \hline
\hhline{|===|}
\rowcolor{green!8}
\multicolumn{3}{|c|}{\rule{0pt}{1.3em}\textit{\textbf{Perception-centric}}\rule[-0.8em]{0pt}{0pt}}                                                                                                                                                                                                                                                                                                                                                                                          \\ 
\hhline{|===|}
% \hline
\multicolumn{3}{|c|}{\rule{0pt}{1.1em} \textbf{Appearance Compliance Generation} \rule[-0.5em]{0pt}{0pt}}                                                                                                                                                                                                                                                                                                                                                                                     \\ \hline
\multicolumn{1}{|c|}{5.1}              & \multicolumn{1}{c|}{\multirow{5}{*}{\begin{tabular}[c]{@{}c@{}}Text \\ Rendering\end{tabular}}}                                         & \multicolumn{1}{l|}{Typeface Rendering}                                                                                                                                                                                                                          \\ \cline{1-1} \cline{3-3} 
\multicolumn{1}{|c|}{5.2}              & \multicolumn{1}{c|}{}                                                                                                                   & \multicolumn{1}{l|}{STEM Formula Generation}                                                                                                                                                                                                                     \\ \cline{1-1} \cline{3-3} 
\multicolumn{1}{|c|}{5.3}              & \multicolumn{1}{c|}{}                                                                                                                   & \multicolumn{1}{l|}{Academic Document Generation}                                                                                                                                                                                                                \\ \cline{1-1} \cline{3-3} 
\multicolumn{1}{|c|}{5.4}              & \multicolumn{1}{c|}{}                                                                                                                   & \multicolumn{1}{l|}{Scene Text Generation in Multiple Languages}                                                                                                                                                                                                 \\ \cline{1-1} \cline{3-3} 
\multicolumn{1}{|c|}{5.5}              & \multicolumn{1}{c|}{}                                                                                                                   & \multicolumn{1}{l|}{Poster Generation in Multiple Languages}                                                                                                                                                                                                     \\ \hline
\multicolumn{1}{|c|}{5.6}              & \multicolumn{1}{c|}{\multirow{5}{*}{\begin{tabular}[c]{@{}c@{}}Controlled Visuals\\ Generation\end{tabular}}}                           & \multicolumn{1}{l|}{Object Instance Count Control Generation}                                                                                                                                                                                                                                \\ \cline{1-1} \cline{3-3} 
\multicolumn{1}{|c|}{5.7}              & \multicolumn{1}{c|}{}                                                                                                                   & \multicolumn{1}{l|}{Action Cardinality Control Generation}                                                                                                                                                                                                       \\ \cline{1-1} \cline{3-3} 
\multicolumn{1}{|c|}{5.8}              & \multicolumn{1}{c|}{}                                                                                                                   & \multicolumn{1}{l|}{Multi-Object Generation with Attribute Diversity}                                                                                                                                                                                            \\ \cline{1-1} \cline{3-3} 
% \multicolumn{1}{|c|}{5.9}              & \multicolumn{1}{c|}{}                                                                                                                   & \multicolumn{1}{l|}{Object Instance Count Control Generation}                                                                                                                                                                                                    \\ \cline{1-1} \cline{3-3} 
\multicolumn{1}{|c|}{5.9}             & \multicolumn{1}{c|}{}                                                                                                                   & \multicolumn{1}{l|}{Relational Spatial Positioning Composition}                                                                                                                                                                                                  \\ \cline{1-1} \cline{3-3} 
\multicolumn{1}{|c|}{5.10}             & \multicolumn{1}{c|}{}                                                                                                                   & \multicolumn{1}{l|}{Absolute Spatial Positioning Generation}                                                                                                                                                                                                     \\ \hline
\multicolumn{3}{|c|}{\rule{0pt}{1.1em} \textbf{Dynamics Consistency Generation} \rule[-0.5em]{0pt}{0pt}}                                                                                                                                                                                                                                                                                                                                                                                      \\ \hline
\multicolumn{1}{|c|}{6.1}              & \multicolumn{1}{c|}{\multirow{1}{*}{\begin{tabular}[c]{@{}c@{}}\end{tabular}}}                               & \multicolumn{1}{l|}{Virtual Try-on}                                                                                                                                                                                                                              \\ \cline{1-1} \cline{3-3} \hline
\multicolumn{1}{|c|}{6.2}              & \multicolumn{1}{c|}{\multirow{3}{*}{\begin{tabular}[c]{@{}c@{}}Subject-driven\\ Generation\end{tabular}}}                                                                                                                   & \multicolumn{1}{l|}{Interior Layout Generation}                                                                                                                                                                                                                  \\ \cline{1-1} \cline{3-3} 
\multicolumn{1}{|c|}{6.3}              & \multicolumn{1}{c|}{}                                                                                                                   & \multicolumn{1}{l|}{Human Portrait Generation}                                                                                                                                                                                                                   \\ \cline{1-1} \cline{3-3} 
\multicolumn{1}{|c|}{6.4}              & \multicolumn{1}{c|}{}                                                                                                                   & \multicolumn{1}{l|}{Multi-Concept Customization}                                                                                                                                                                                                                 \\ \hline
\multicolumn{1}{|c|}{6.5}              & \multicolumn{1}{c|}{\begin{tabular}[c]{@{}c@{}}Story\\ Generation\end{tabular}}                                                         & \multicolumn{1}{l|}{Narrative-to-Comic Generation}                                                                                                                                                                                                               \\ \hline
\multicolumn{1}{|c|}{6.6}              & \multicolumn{1}{c|}{\multirow{3}{*}{\begin{tabular}[c]{@{}c@{}}Fine-grained\\ Image Editing\end{tabular}}}                              & \multicolumn{1}{l|}{Fine-grained Image Editing – Addition}                                                                                                                                                                                                       \\ \cline{1-1} \cline{3-3} 
\multicolumn{1}{|c|}{6.7}              & \multicolumn{1}{c|}{}                                                                                                                   & \multicolumn{1}{l|}{Fine-grained Image Editing – Removal}                                                                                                                                                                                                        \\ \cline{1-1} \cline{3-3} 
\multicolumn{1}{|c|}{6.8}              & \multicolumn{1}{c|}{}                                                                                                                   & \multicolumn{1}{l|}{Fine-grained Image Editing – Replacement}                                                                                                                                                                                                    \\ \hline
                                       % & \multicolumn{1}{l}{}                                                                                                                    &                                                                                                                                                                                                                                                                 
\end{longtable}

% wk1
\begin{figure*}[t]
    \centering
    \includegraphics[width=1.0\textwidth]{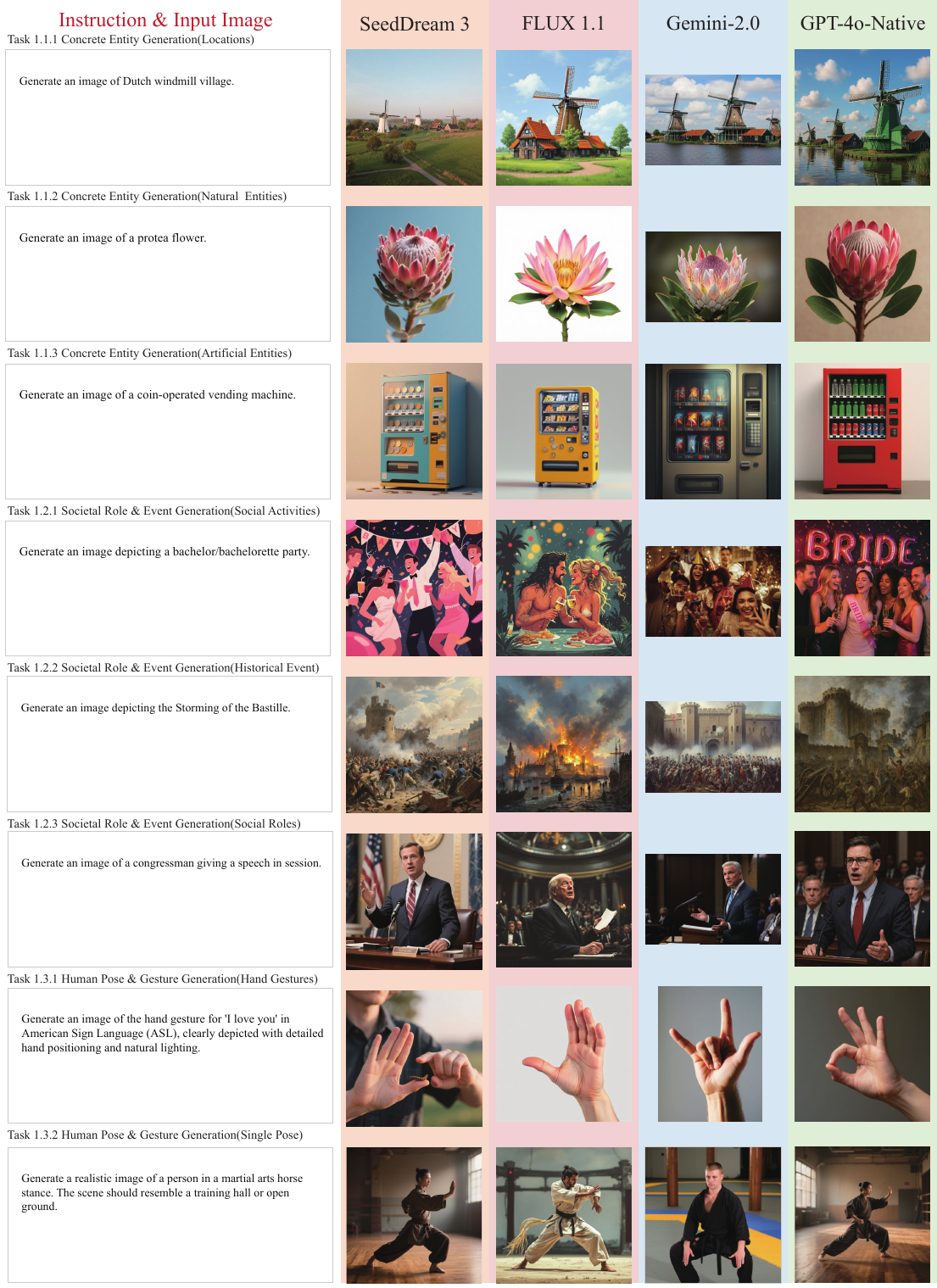}
    \caption{Examples of tasks from the \textbf{World Knowledge Anchored Generation} section, along with the corresponding outputs generated by different models for qualitative comparison.}
    \label{wk1}
\end{figure*}

% wk2
\begin{figure*}[t]
    \centering
    \includegraphics[width=1.0\textwidth]{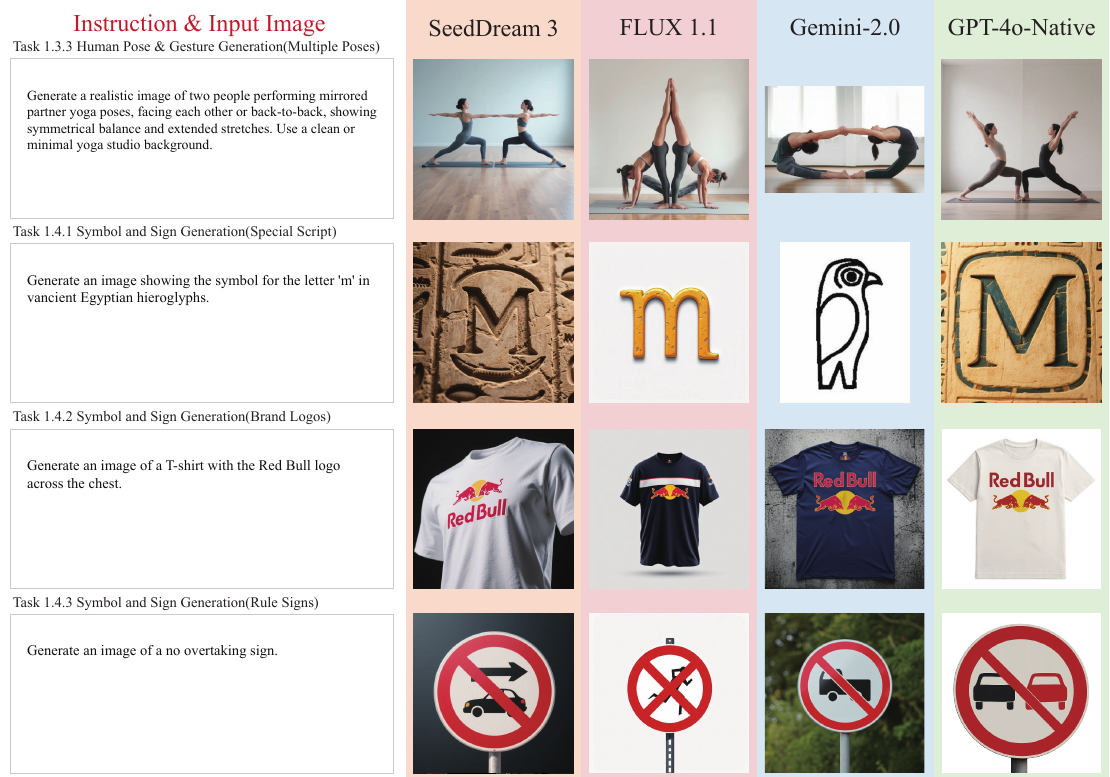}
    \caption{Examples of tasks from the \textbf{World Knowledge Anchored Generation} section, along with the corresponding outputs generated by different models for qualitative comparison.}
    \label{wk2}
\end{figure*}

% sit1
\begin{figure*}[t]
    \centering
    \includegraphics[width=1.0\textwidth]{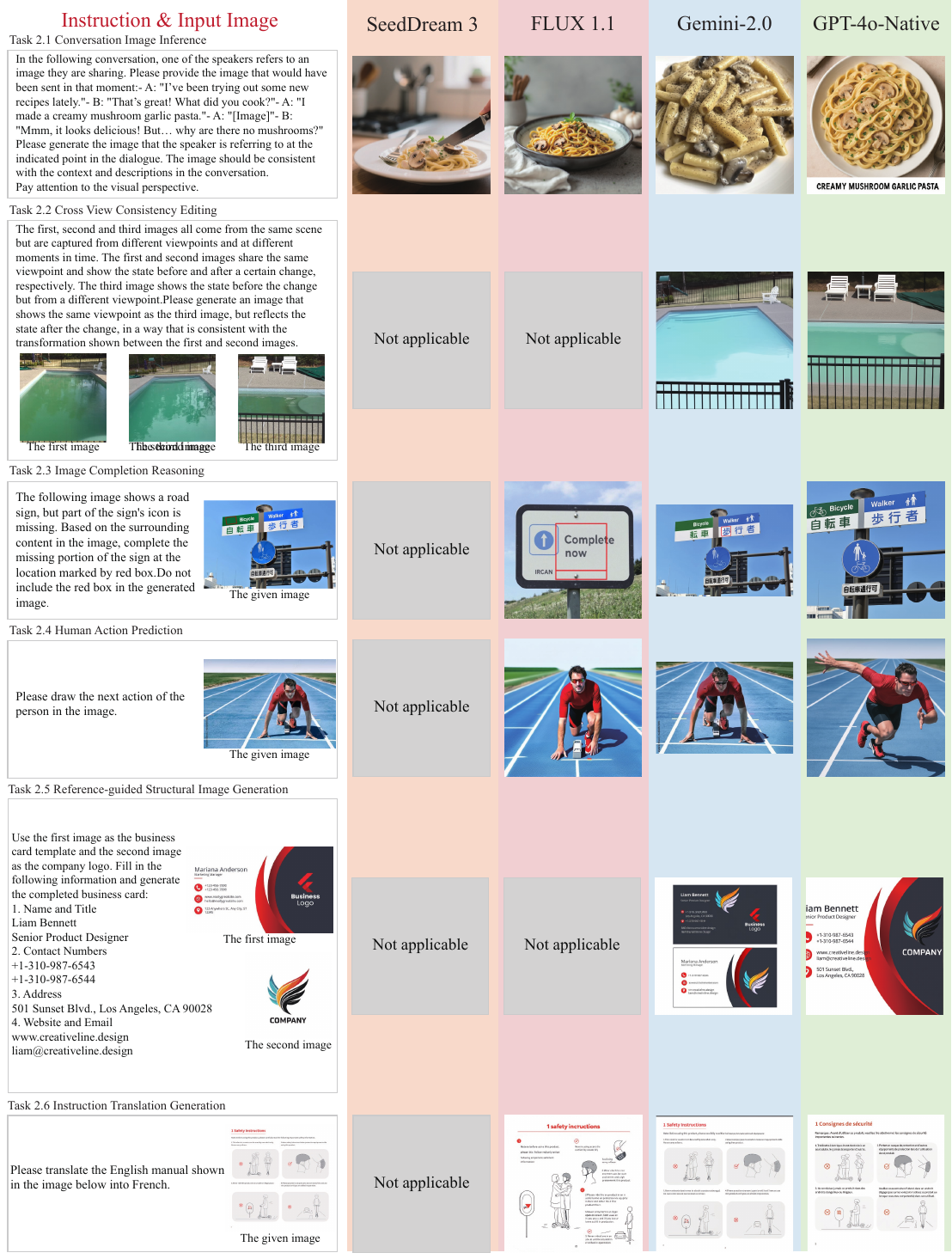}
    \caption{Examples of tasks from the \textbf{Situational Reasoning Generation} section, along with the corresponding outputs generated by different models for qualitative comparison.}
    \label{sit1}
\end{figure*}

% sit2
\begin{figure*}[t]
    \centering
    \includegraphics[width=1.0\textwidth]{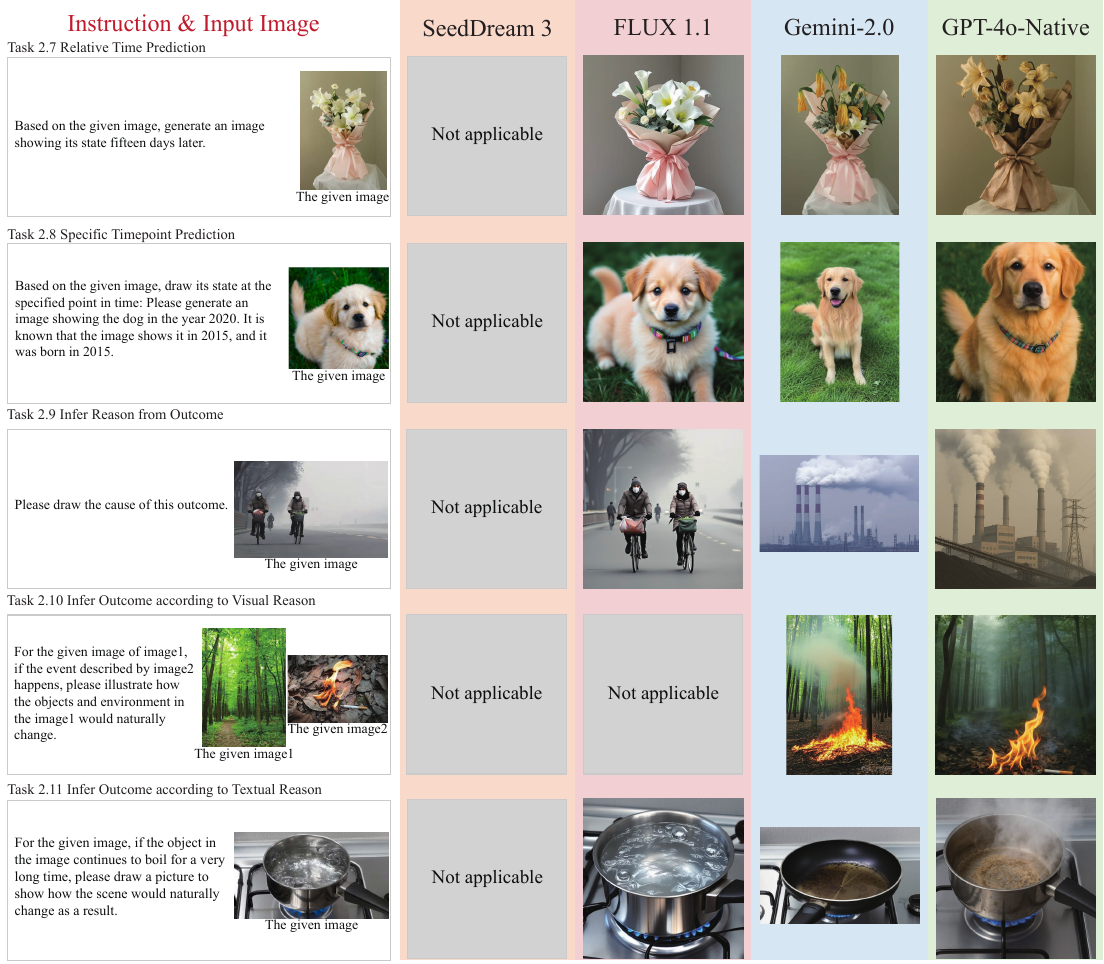}
    \caption{Examples of tasks from the \textbf{Situational Reasoning Generation} section, along with the corresponding outputs generated by different models for qualitative comparison.}
    \label{sit2}
\end{figure*}

% spa1
\begin{figure*}[t]
    \centering
    \includegraphics[width=1.0\textwidth]{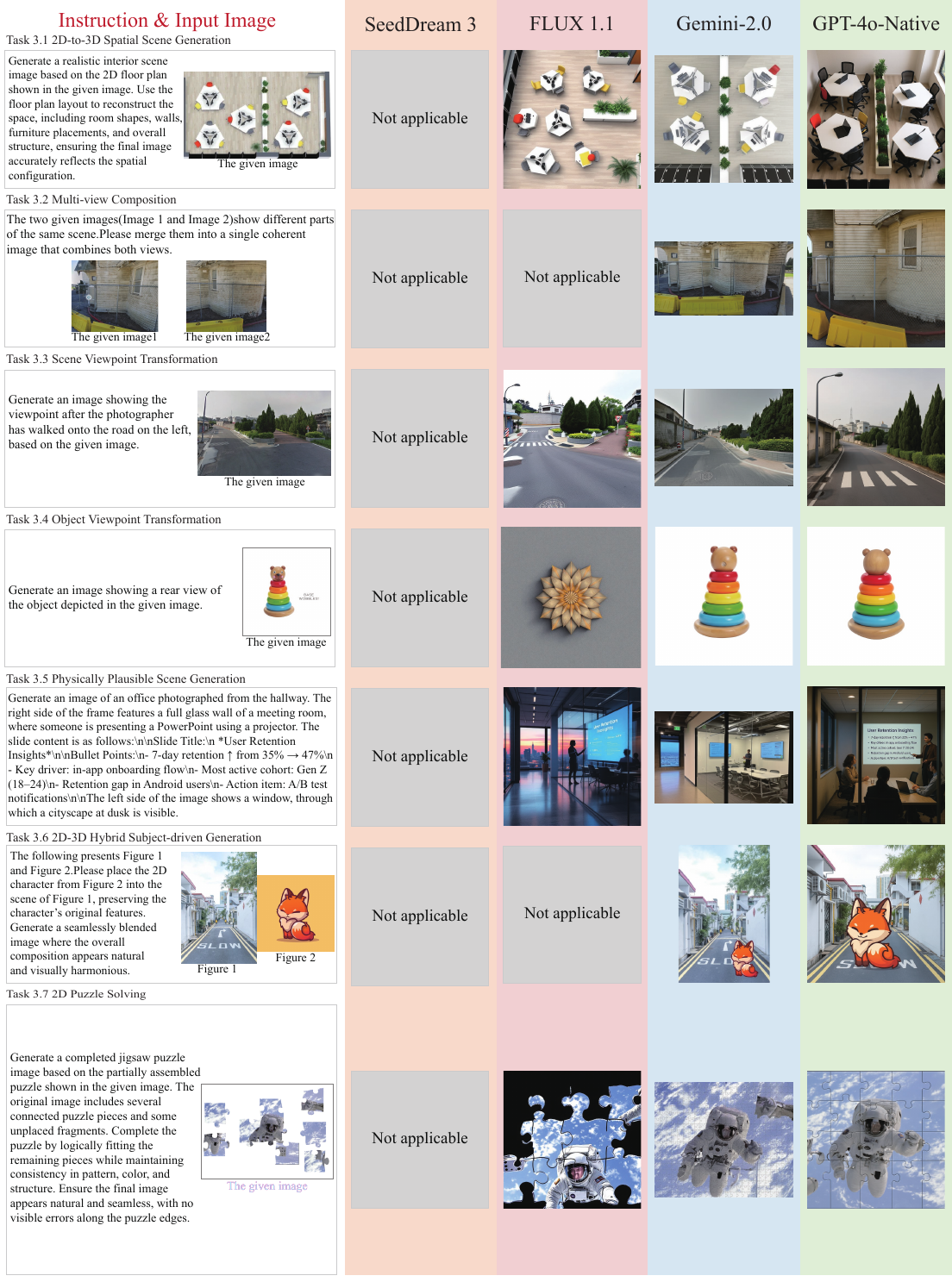}
    \caption{Examples of tasks from the \textbf{Spatial Reasoning Generation} section, along with the corresponding outputs generated by different models for qualitative comparison.}
    \label{spa1}
\end{figure*}

% spa2
\begin{figure*}[t]
    \centering
    \includegraphics[width=1.0\textwidth]{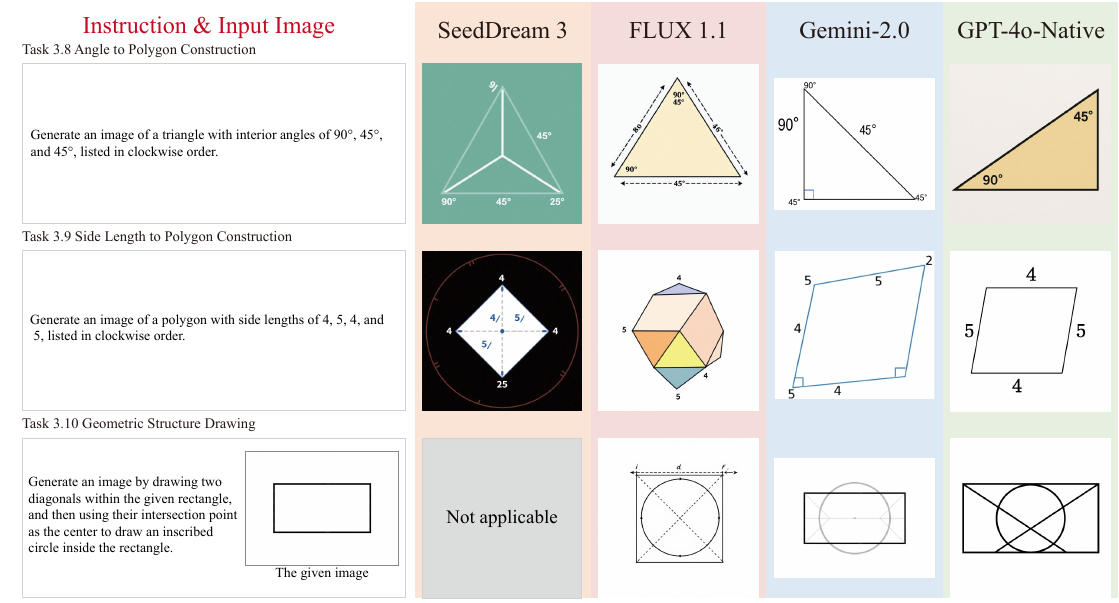}
    \caption{Examples of tasks from the \textbf{Spatial Reasoning Generation} section, along with the corresponding outputs generated by different models for qualitative comparison.}
    \label{spa2}
\end{figure*}

% stem1
\begin{figure*}[t]
    \centering
    \includegraphics[width=1.0\textwidth]{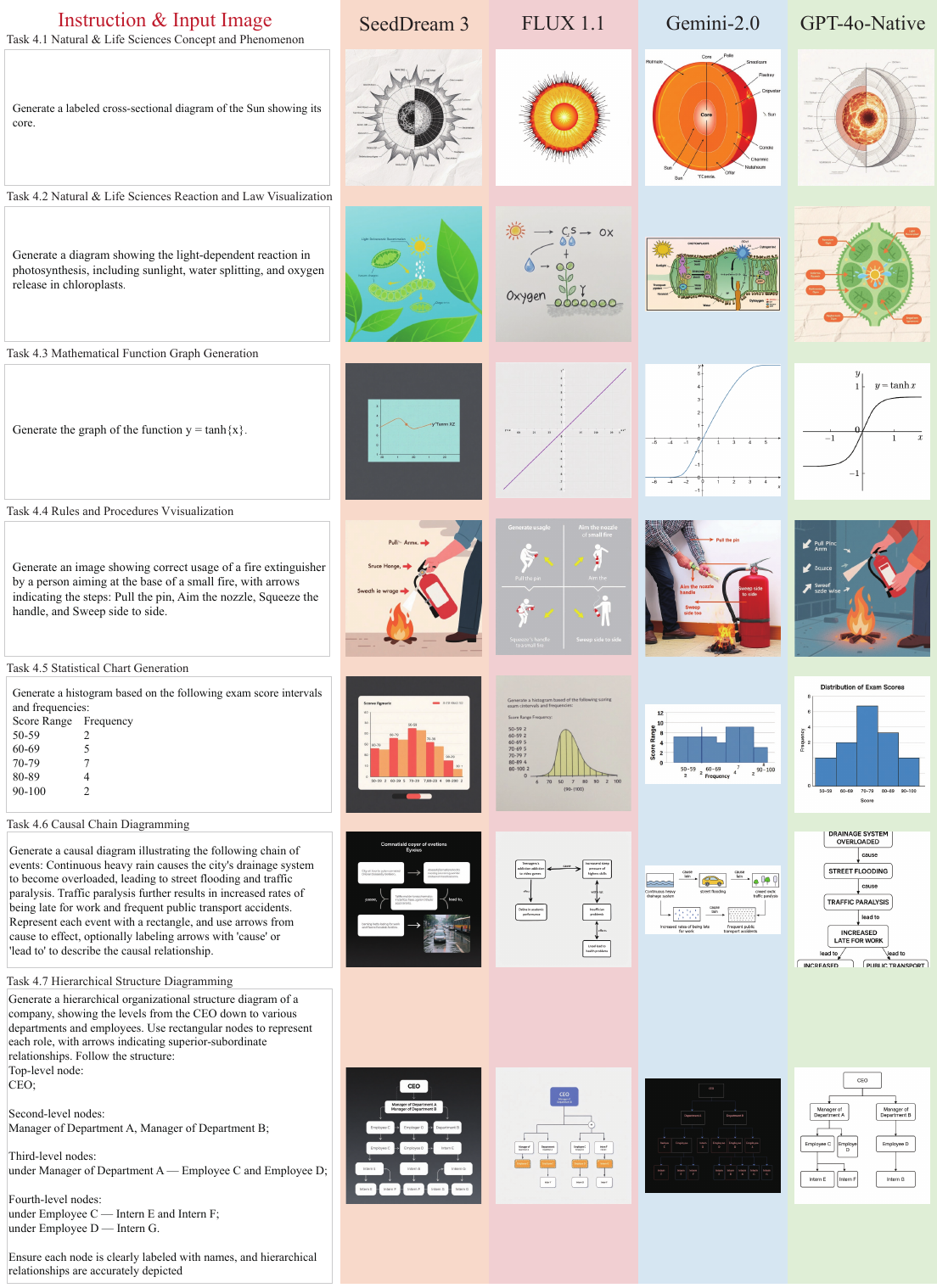}
    \caption{Examples of tasks from the \textbf{STEM-Driven Reasoning Generation} section, along with the corresponding outputs generated by different models for qualitative comparison.}
    \label{stem1}
\end{figure*}

% stem2
\begin{figure*}[t]
    \centering
    \includegraphics[width=1.0\textwidth]{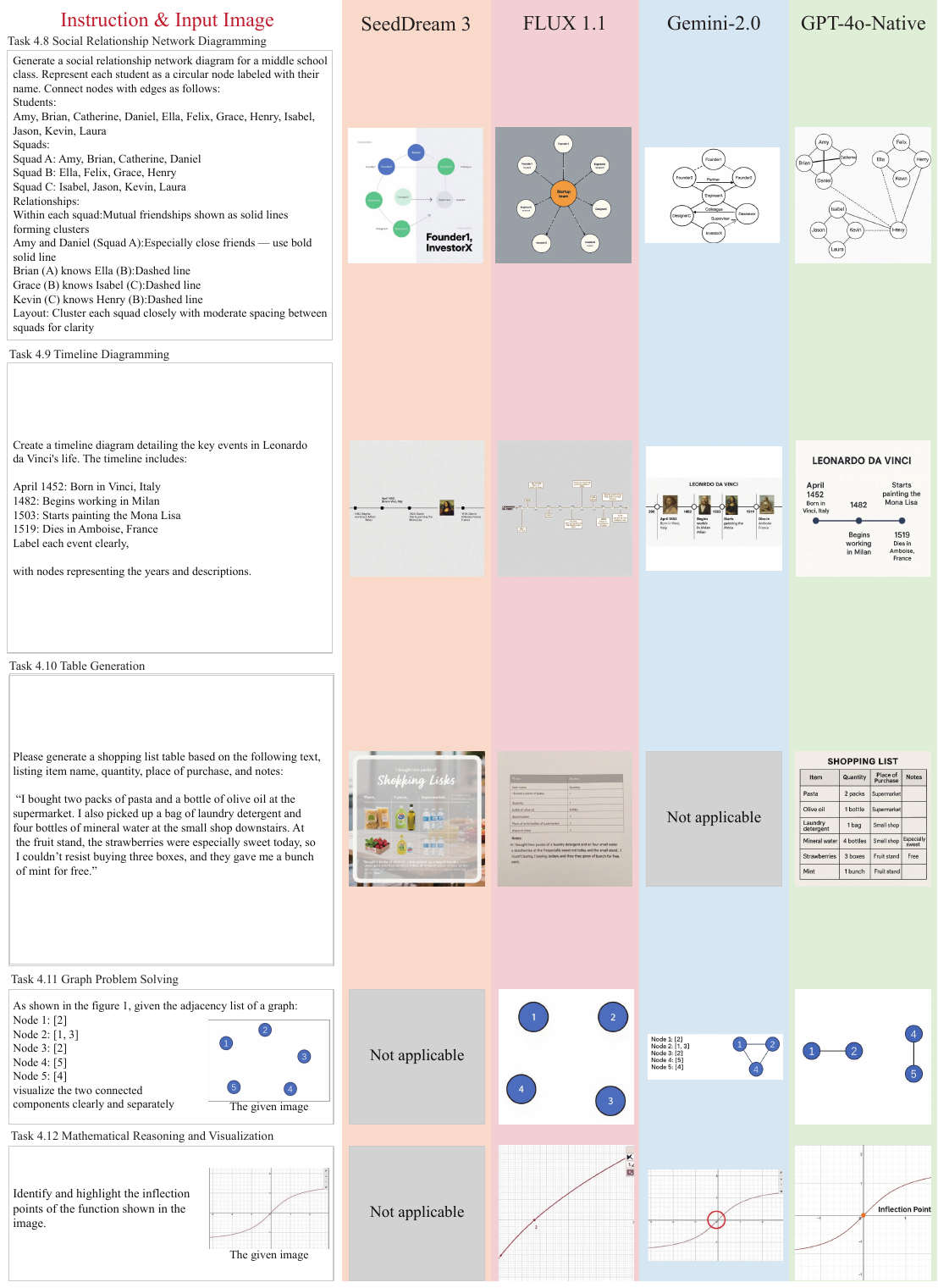}
    \caption{Examples of tasks from the \textbf{STEM-Driven Reasoning Generation} section, along with the corresponding outputs generated by different models for qualitative comparison.}
    \label{stem2}
\end{figure*}

% stem3
\begin{figure*}[t]
    \centering
    \includegraphics[width=1.0\textwidth]{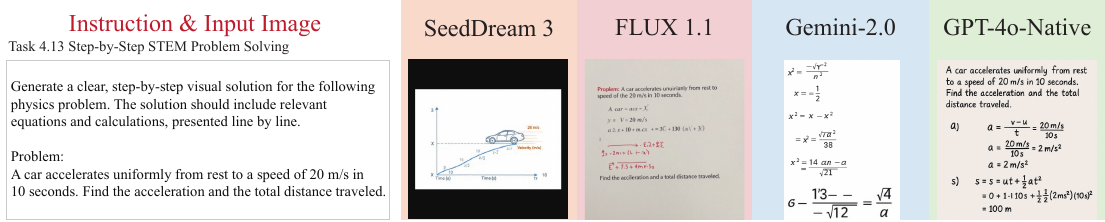}
    \caption{Examples of tasks from the \textbf{STEM-Driven Reasoning Generation} section, along with the corresponding outputs generated by different models for qualitative comparison.}
    \label{stem3}
\end{figure*}

% app1
\begin{figure*}[t]
    \centering
    \includegraphics[width=1.0\textwidth]{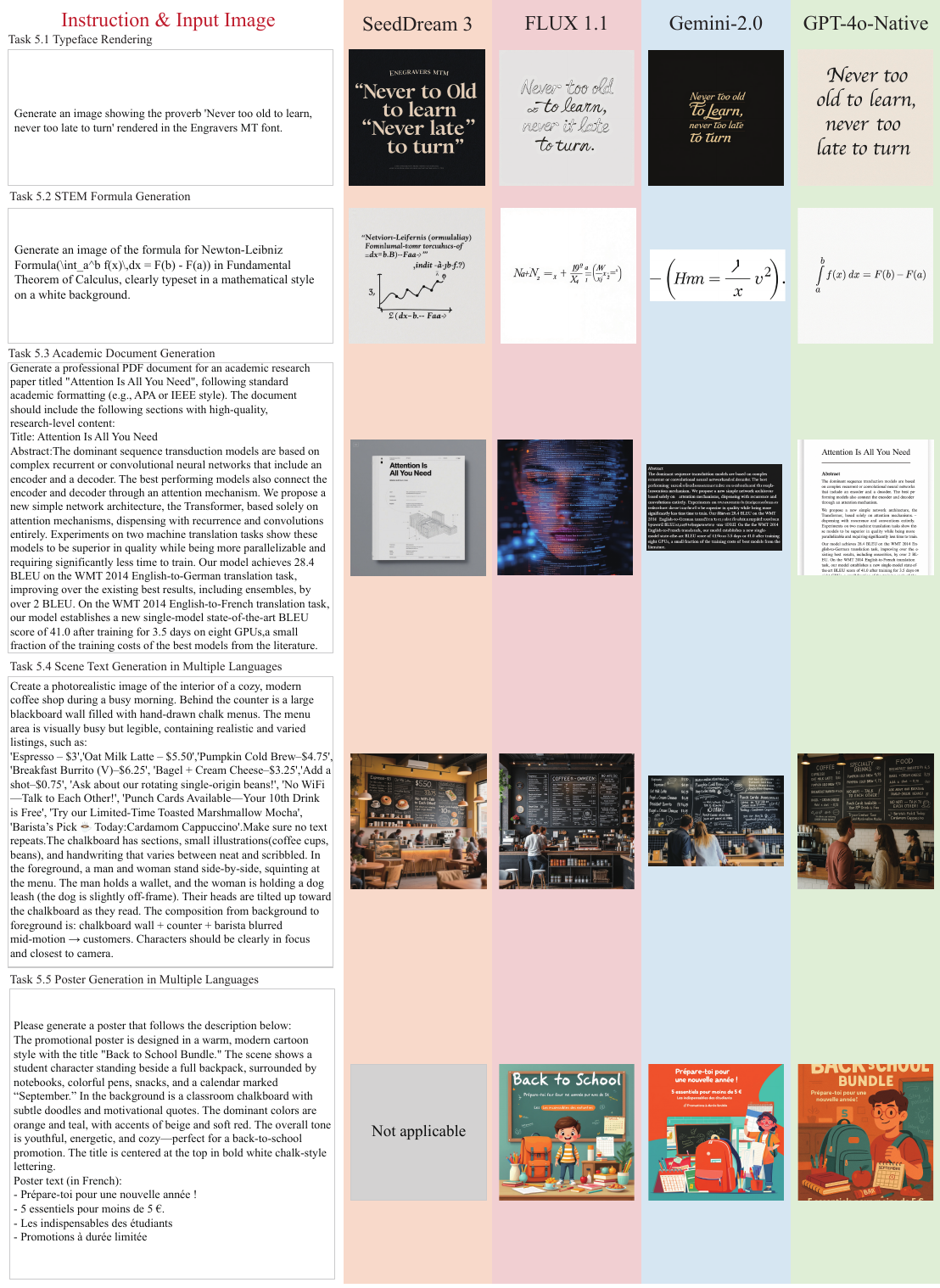}
    \caption{Examples of tasks from the \textbf{Appearance Compliance Generation} section, along with the corresponding outputs generated by different models for qualitative comparison.}
    \label{app1}
\end{figure*}

% app2
\begin{figure*}[t]
    \centering
    \includegraphics[width=1.0\textwidth]{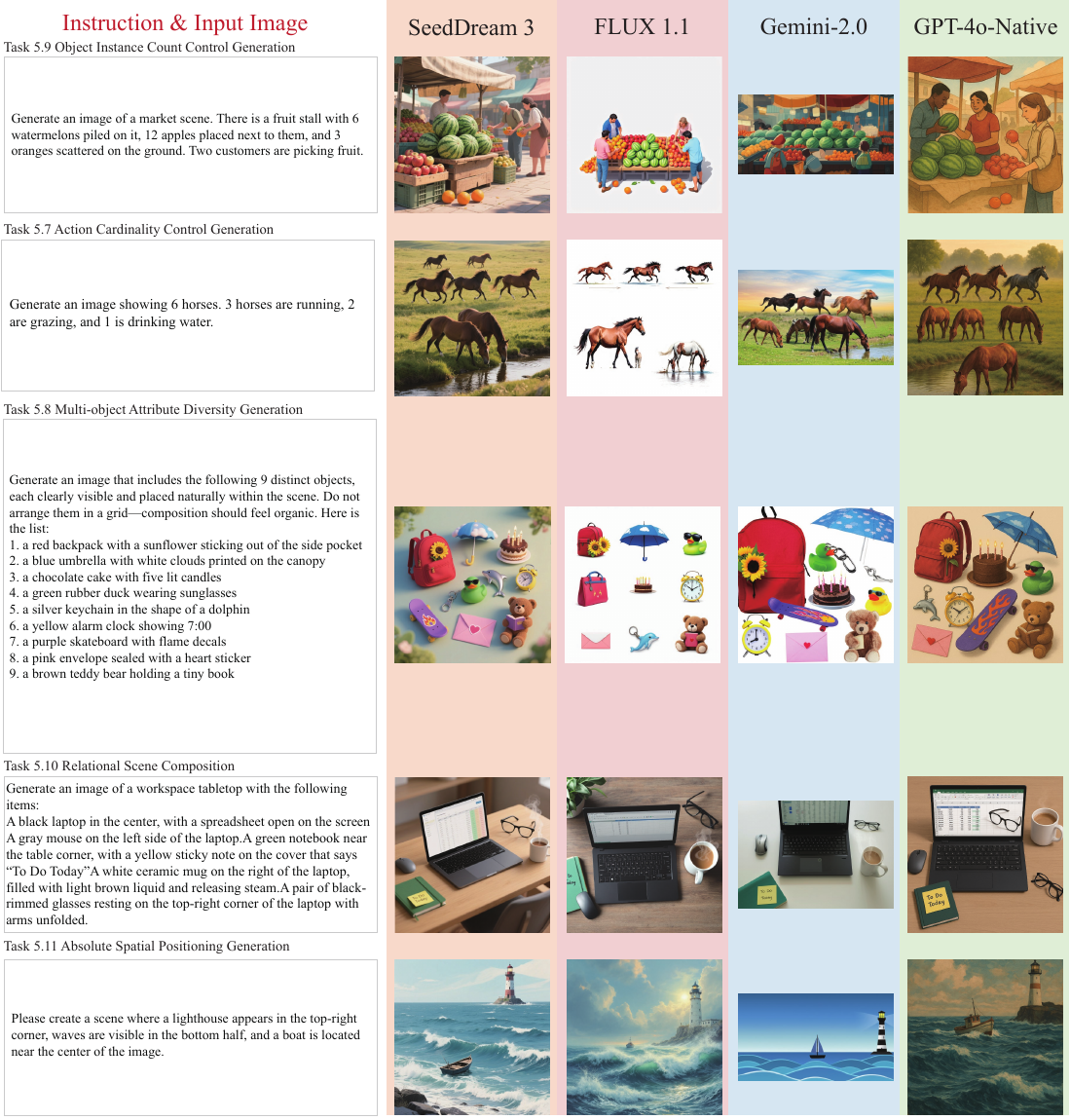}
    \caption{Examples of tasks from the \textbf{Appearance Compliance Generation} section, along with the corresponding outputs generated by different models for qualitative comparison.}
    \label{app2}
\end{figure*}

% dyn1
\begin{figure*}[t]
    \centering
    \includegraphics[width=1.0\textwidth]{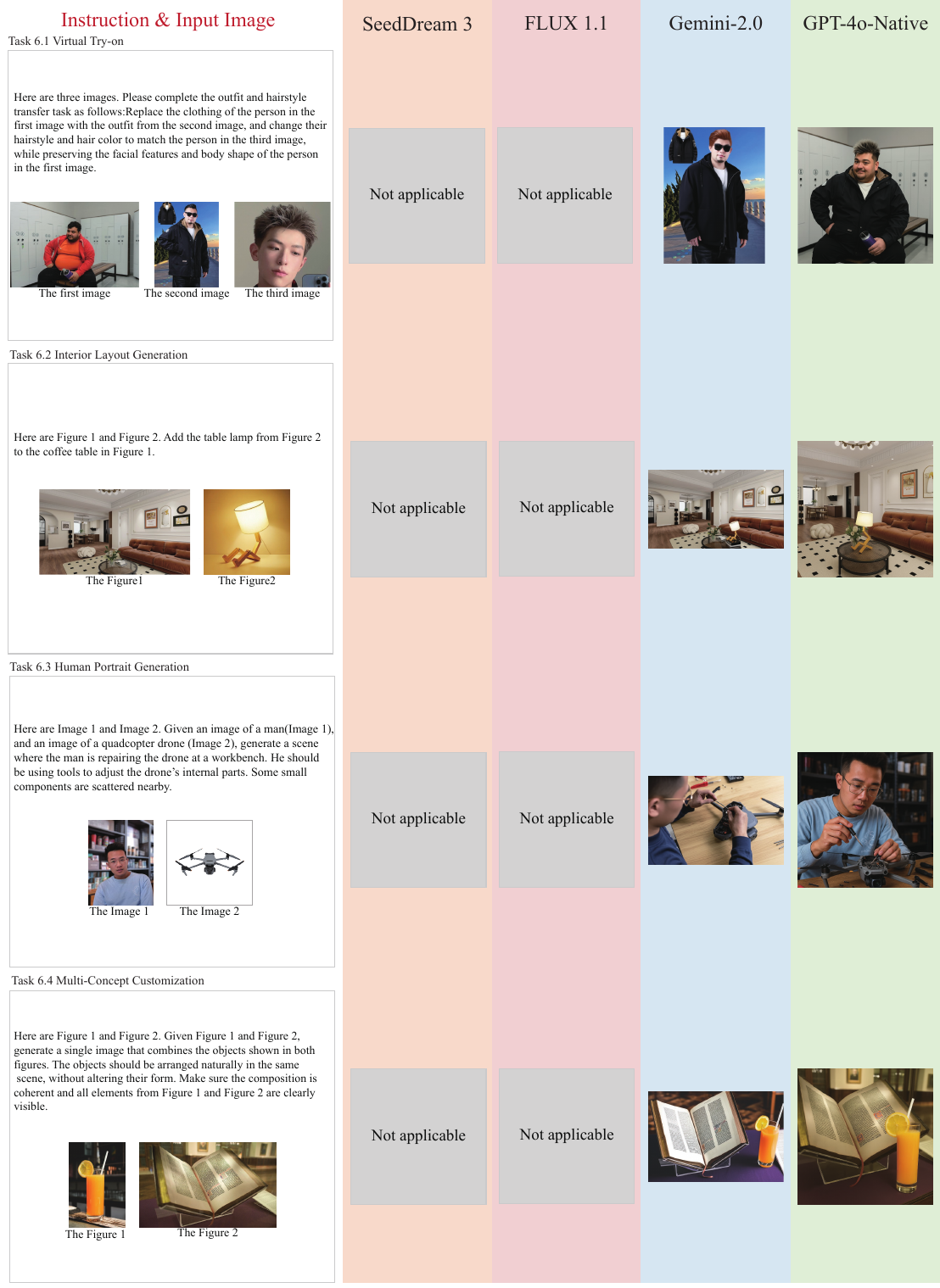}
    \caption{Examples of tasks from the \textbf{Dynamics Consistency Generation} section, along with the corresponding outputs generated by different models for qualitative comparison.}
    \label{dyn1}
\end{figure*}

% dyn2
\begin{figure*}[t]
    \centering
    \includegraphics[width=1.0\textwidth]{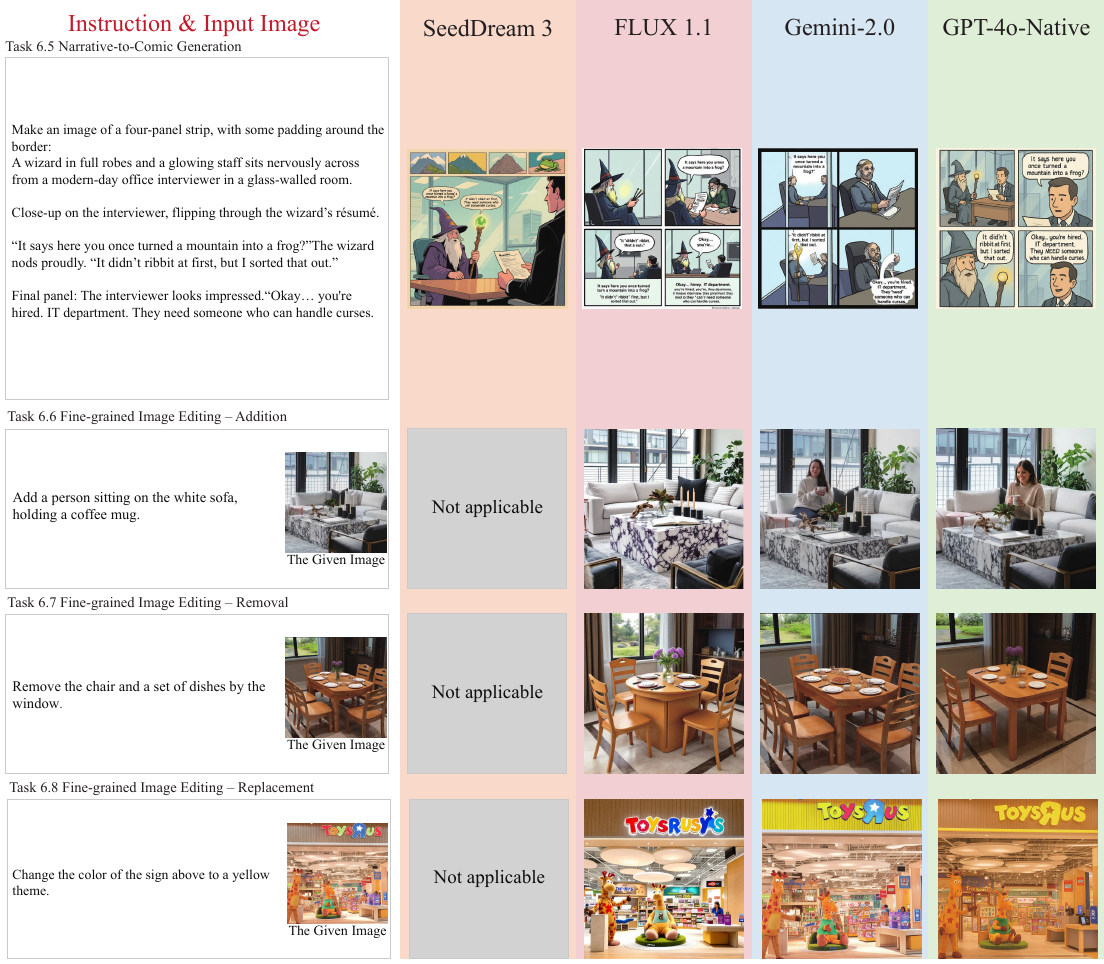}
    \caption{Examples of tasks from the \textbf{Dynamics Consistency Generation} section, along with the corresponding outputs generated by different models for qualitative comparison.}
    \label{dyn2}
\end{figure*}

\end{document}